\documentclass[12pt,journal,draftclsnofoot,onecolumn]{IEEEtran}

\usepackage{mathptmx}   
\usepackage{amsfonts, amssymb, amsmath}
\usepackage{epsfig}
\usepackage{theorem}
\usepackage{textcomp}
\usepackage{tipa}
\usepackage{multirow}

\makeatletter
\let\NAT@parse\undefined
\makeatletter
\usepackage{natbib}

\usepackage[linesnumbered,ruled]{algorithm2e}
\SetKwInput{KwIn}{Initial condition}
\SetKwInput{KwResult}{Outcome}
\SetAlgorithmName{Strategy}{List of strategies}

\newtheorem{theorem}{Theorem}

\newtheorem{corollary}[theorem]{Corollary}
\newtheorem{lemma}[theorem]{Lemma}
\newtheorem{problem}{Problem}

\newcommand{\qed}{\hfill $\Box$\\}

\makeatletter
\newcommand{\customlabel}[2]{%
\protected@write \@auxout {}{\string \newlabel {#1}{{#2}{}}}}
\makeatother

\begin{document}

\title{Target Assignment in Robotic Networks: Distance Optimality Guarantees and Hierarchical Strategies}
\author{
\begin{tabular}{c}
Jingjin~Yu,~Soon-Jo~Chung,~Petros~G.~Voulgaris
\end{tabular}
\thanks{Jingjin Yu is currently with the Computer Science and Artificial Intelligence Lab at the Massachusetts Institute of Technology. E-mail: jingjin@csail.mit.edu. The work was completed when J. Yu was at the University of Illinois at Urbana-Champaign. Soon-Jo Chung and Petros G. Voulgaris are with the Coordinated Science Lab and the Department of Aerospace Engineering at the University of Illinois at Urbana-Champaign. E-mail: \{sjchung, voulgari\}@illinois.edu. This work was supported in part by AFOSR grant FA95501210193, NSF grant IIS-1253758, and NSF grant ECCS 10-27437.} 
}
\maketitle

\begin{abstract}We study the problem of multi-robot target assignment to minimize the total distance traveled by the robots until they all reach an equal number of static targets. In the first half of the paper, we present a necessary and sufficient condition under which true distance optimality can be achieved for robots with limited communication and target-sensing ranges. Moreover, we provide an explicit, non-asymptotic formula for computing the number of robots needed to achieve distance optimality in terms of the robots' communication and target-sensing ranges with arbitrary guaranteed probabilities. The same bounds are also shown to be asymptotically tight. 

In the second half of the paper, we present suboptimal strategies for use when the number of robots cannot be chosen freely. Assuming first that all targets are known to all robots, we employ a hierarchical communication model in which robots communicate only with other robots in the same partitioned region. This hierarchical communication model leads to constant approximations of true distance-optimal solutions under mild assumptions. We then revisit the limited communication and sensing models. By combining simple rendezvous-based strategies with a hierarchical communication model, we obtain decentralized hierarchical strategies that achieve constant approximation ratios with respect to true distance optimality. Results of simulation show that the approximation ratio is as low as 1.4. 

\end{abstract}

\section{Introduction}
\IEEEPARstart{I}n this paper, we study the permutation-invariant assignment of a set of networked robots to a set of targets of equal cardinality. Focusing on minimizing the total distance traveled by the robots in a planar setting, we seek optimality guarantees and near-optimal strategies. For robot-to-robot communication, we investigate both a simple circular range-based model and a region-based model in which all robots within the same region can communicate with each other. When we consider the limited target-sensing capability of the robots, a circular range sensing model is used.

The class of problems that we study is denoted as {\em target assignment in robotic networks} as it shares many similarities with the problems studied in \cite{SmiBul09}. In \cite{SmiBul09}, the authors characterized the performance of {\sc ETSP\footnote{{\sc ETSP} stands for the {\em Euclidean traveling salesman problem}.} Assgmt} and {\sc GRID Assgmt} algorithms (strategies) in achieving time optimality ({\em i.e.}, minimizing the time until every target is occupied). In contrast, we focus on minimizing the total distance traveled by all robots with significantly different assumptions on the communication and sensing models of the robots. The total distance serves as a proper proxy to quantities such as the total energy consumption of all the robots. Note that a distance-optimal solution for the target assignment problem generally does not imply time optimality and vice versa \cite{YuLav12CDC}. 

As its name implies, the problem of {\em target assignment in robotic networks} requires solving an {\em assignment} (or {\em matching}) problem. The assignment problem is extensively studied in the area of combinatorial optimization, with polynomial time algorithms available for solving many of its variations \cite{Ber88,BerCas91,BurDelMar12,EdmKar72,Kuh55,ZavSpePap08}. If we instead put more emphasis on multi-robot systems, the problems of robotic task allocation \cite{JiAzuEge06,TanJadPap07,TrePavFra13,ZavPap08}, swarm reconfiguration \cite{ChuBanChaHad13}, multi-robot path planning \cite{KloHut06,ShaSavFraVou07,TurMohMicKum13}, and multi-agent consensus \cite{CorMarBul06,JadLinMor03,LinMorAnd07A,LinMorAnd07B} are relevant. For a more comprehensive review on these topics, see \cite{BulCorMar09}. 

Our work is also closely related to the study of the connectivity of wireless networks. An interesting result \cite{XueKum04} showed that, if $n$ robots are uniformly randomly scattered in a unit square, then each robot needs to communicate with $k = \Theta(\log n)$ nearest neighbors for the entire robotic network to be asymptotically connected as $n$ approaches infinity. In particular, the authors of \cite{XueKum04} showed that $k < 0.074\log n$ leads to an asymptotically disconnected network whereas $k > 5.1774 \log n$ guarantees  asymptotic connectivity. This pair of bounds was subsequently improved and extended in \cite{BBSW05}. These nearest neighbor based connectivity models were further studied in \cite{FreKowKum10,GanXue07,MaoAnd13}, to list a few. In many of these settings, a {\em geometric graph} structure is used \cite{Pen03}.

This research effort brings forth three contributions. First, for robots with limited range-based target-sensing and communication capabilities (the ranges are captured by radii $r_{\mathrm{sense}}$ and $r_{\mathrm{comm}}$, respectively), we derive necessary and sufficient conditions for ensuring a distance-optimal solution. In particular, we provide a probabilistic estimate of the number of robots (denoted by $n$) sufficient for all robots to form a connected network for a fixed communication radius $r_{\mathrm{comm}}$. In contrast to the asymptotic connectivity results from \cite{XueKum04,Pen97}, we give $n$ as an explicit function of $r_{\mathrm{comm}}$ that is also non-asymptotic. Therefore, our bounds hold without requiring $n \to \infty$ . We further show that the same bounds are asymptotically tight when a high probability guarantee is required. 

Second, allowing the robots to have global target-sensing capabilities coupled with a region-based communication model, we show that an infinite family of hierarchical strategies can lead to decentralized target assignments while ensuring that the total expected distance traveled by the robots is asymptotically within a constant multiple of the optimal distance. Our simulation results show that this bound can often be smaller than two. Moreover, because hierarchical strategies avoid running a centralized assignment algorithm, significant savings on computation time (in certain cases, a speedup of $1000$ times or more) are achieved. 

Third, for robots with global target-sensing capabilities and a range-based communication model, hierarchical strategies (for assignment) and rendezvous-based strategies (for compensating for the lack of global communication) are combined to obtain decentralized suboptimal algorithms. These hybrid strategies, under mild assumptions, preserve the constant approximation ratios on distance optimality achieved by the ``pure'' hierarchical strategies. We further show that the global target-sensing assumption can be removed without affecting asymptotic optimality. 

Portions of this work were presented in \cite{YuChuVou14ICRA, YuChuVou14ISCCSP} for the early dissemination of results. Compared with \cite{YuChuVou14ICRA, YuChuVou14ISCCSP}, this paper provides a more comprehensive view of the results along with complete proofs for all theorems. Many of the proofs have been significantly improved to illustrate more clearly proof techniques that may be of interest on their own. In addition, the current paper discusses extensively generalizations of the stochastic target assignment problem to mismatching number of robots and targets, and to higher dimensions. 

The rest of the paper is organized as follows. In Section \ref{sec:prelim}, we present notations and well-known results from other branches of research needed for the development of our results. After stating the problem formally in Section \ref{sec:formulation}, we then elaborate on the three stated contributions in Sections \ref{sec:bound}-\ref{sec:strat}. We present results of simulation studies in Section \ref{sec:exp} to validate our theoretical results and conclude in Section \ref{sec:con}.

\section{Preliminaries}\label{sec:prelim}
In this section, we review results on the balls and bins problem, linear assignment, and random geometric graphs. The symbols $\mathbb R, \mathbb R^+, \mathbb N$ denote the set of real numbers, the set of positive reals, and the set of positive integers, respectively. For a positive real number $x$, $\log x$ denotes the natural logarithm of $x$;  the function $\lceil x \rceil$ (resp.,  $\lfloor x \rfloor$) denotes the smallest (resp.,  largest) integer that is larger (resp.,  smaller) than or equal to $x$. $|\cdot|$ denotes the cardinality for a set and the absolute value for a real number. We use $\Vert v \Vert_2$ to denote the Euclidean $2$-norm of a vector $v$. The unit square $[0, 1]^2 \subset \mathbb R^2$ is denoted as $Q$. The expectation of a random variable $X$ is denoted as $\mathbf{E}[X]$. We use $E(\cdot)$ to represent a probabilistic event and the probability with which an event $e$ occurs is denoted as $\mathbf{P}(e)$. 

Given two functions $f, g: \mathbb R^+ \to \mathbb R^+$, $f(x) = O(g(x))$ if and only if there exist $M_O, x_O \in \mathbb R^+$ such that 
\begin{displaymath}
\forall x > x_O, |f(x)| \le M_O|g(x)|.
\end{displaymath}

Similarly, $f(x) = \Omega(g(x))$ if and only if there exist $M_{\Omega}, x_{\Omega} \in \mathbb R^+$ such that 
\begin{displaymath}
\forall x > x_{\Omega}, |f(x)| \ge M_{\Omega}|g(x)|.
\end{displaymath}

If $f(x) = O(g(x))$ and $f(x) = \Omega(g(x))$, then we say $f(x) = \Theta(g(x))$. Finally, $f(x) = o(g(x))$ (resp.,  $f(x) = \omega(g(x))$) if and only if $f(x) = O(g(x))$ (resp.,  $f(x) = \Omega(g(x))$) and $f(x) = \Theta(g(x))$ does not hold. 

\subsection{Balls and Bins}
The well-studied problem in probability theory known as the {\em urns-problem}, or the problem of {\em balls and bins}, considers the distribution generated as a number of balls are randomly tossed into a set of bins. The following classical result on the ball and bins problem is due to Erd\H{o}s and R\'{e}nyi.

\begin{theorem}[Balls and Bins \cite{ErdRen61}]\label{t:er} Suppose that a number of balls are tossed uniformly randomly into $m$ bins, one ball per time step. Let $T_k$ denote the first time such that $k$ balls are collected in every bin. Then for any real number $c$, 
\begin{equation}\label{eq:er}
\lim_{m\to\infty}\mathbf{P}(T_k < m \log m + (k-1)m\,\log\log\,m + cm) = \displaystyle e^{-e^{-\frac{c}{(k-1)!}}}. 
\end{equation}
\end{theorem}

It is worth noting that the proof of Theorem \ref{t:er} is fairly short and elegant, employing only basic tools from analysis and combinatorics. A useful corollary for $k = 1$ follows readily. 

\begin{corollary}\label{c:er1} For an arbitrary real number $c$, suppose that no fewer than $(m \log m + cm)$ balls are tossed uniformly randomly into $m$ bins. As $m \to \infty$, every bin contains at least one ball with probability $e^{-e^{-c}}$. 
\end{corollary}
\noindent {\sc Proof.} In (\ref{eq:er}), letting $k = 1$ yields
\begin{equation}\label{eq:er1}
\lim_{m\to\infty}\mathbf{P}(T_1 < m \log m + cm) = \displaystyle e^{-e^{-c}}. 
\end{equation}
The corollary directly follows (\ref{eq:er1}) (recall that $T_1$ is the number of tosses needed so that every bin has at least one ball). 
~\qed

Corollary \ref{c:er1} says that $T_1 = m\log m$ is a {\em sharp} threshold. Letting $c = 5$ in (\ref{eq:er1}) yields that the probability of every bin being occupied by at least one ball is greater than $0.99$ when at least $m\log m + 5m$ balls are tossed. On the other hand, the same probability is no more than $0.001$ when no more than $m\log m - 2m$ balls are tossed. 

\subsection{Linear Assignment Problem}
The {\em (linear) assignment problem}, as a fundamental combinatorial optimization problem, can be defined as follows. 
\begin{problem}[Linear Assignment]\label{p:lap}
Given two finite sets $X$ and $Y$ with $|X| = |Y|$, together with a weight function $C: X\times Y \to \mathbb R$, find a bijection $f: X \to Y$ that minimizes the cost
\begin{equation}\label{eq:lap}
\sum_{x \in X} C(x, f(x)).
\end{equation}
\end{problem}

Problem \ref{p:lap} is also called the {\em perfect weighted bipartite matching} problem. Here, the mapping $C$ is essentially a square matrix that can be used to represent a variety of weights, such as the Euclidean distance when $X$ and $Y$ represent physical locations. The {\em Hungarian method} for the assignment problem, proposed by Kuhn \cite{Kuh55}, has an $O(n^4)$ running time, which was subsequently improved to $O(n^3)$ by Edmonds and Karp \cite{EdmKar72}. Many other algorithms for the assignment problem exist, including other primal-dual (linear programming) methods \cite{BurDelMar12}, auction based methods \cite{Ber88}, and parallel algorithms \cite{BerCas91,ZavSpePap08}. Nevertheless, the strongly polynomial\footnote{A polynomial time algorithm runs in {\em strongly polynomial time} only if its running time does not depend on the {\em size} of the input parameters. Note that $n$ is the {\em number} of input parameters in this case.} $O(n^3)$ Hungarian method remains as the fastest exact (sequential) algorithm, which we use in our simulations.

When $X$ and $Y$ are restricted to points on the plane with the weight function $C$ being the Euclidean distances between the points, the special linear assignment problem is known as the {\em Euclidean bipartite matching} problem, which can be solved exactly using an $O(n^{2.5}\log n)$ primal-dual algorithm \cite{Vai89}. Alternatively, near linear time approximation algorithms can yield near optimal solutions with high probability \cite{ShaAga12}.\footnote{Although algorithms from \cite{ShaAga12,Vai89} have theoretically faster running times than the Hungarian method and apply to the problem that we study, they are more difficult to implement and slower in practice unless $|X|$ is very large because they are not strongly polynomial time algorithms like the Hungarian method.} 

\subsection{Random Geometric Graphs}

Let $X = \{x_1, \ldots, x_n\}$ be a set of $n$ points in the unit square $Q$. For a fixed {\em communication radius} $r_{\mathrm{comm}}$, the {\em geometric graph} $G$ over this set of points is formed by taking each point as a vertex and connecting any two vertices whose underlying points $x_1$ and $x_2$ satisfy $\Vert x_1 - x_2\Vert_2 \le r_{\mathrm{comm}}$. When $X$ is selected randomly following some distribution, the resulting graph is called a {\em random geometric graph}. 

Properties of random geometric graphs have been studied extensively; see \cite{Pen03} for a thorough coverage. One such property is the connectivity of these graphs, which is of particular interest to wireless communication and robotic networks. 
\begin{theorem}[Random Geometric Graphs \cite{Pen97}]\label{t:rgg} Let $G$ be a random geometric graph obtained following the uniform distribution over the unit square for some $n$ and $r_{\mathrm{comm}}$. Then for any real number $c$, as $n \to \infty$, 
\begin{equation}\label{eq:rgg}
\mathbf{P}(\textrm{$G$ is connected} \mid \pi nr_{\mathrm{comm}}^2 - \log n \le c) = e^{-e^{-c}}.
\end{equation}
\end{theorem}

From (\ref{eq:rgg}), it is possible to estimate the number of robots required to guarantee a connected geometric graph $G$. 

\section{Target Assignment in Robotic Networks}\label{sec:formulation}
In this section, we formally define the problem of {\em target assignment in robotic networks} and the optimality objective. 
\subsection{Problem Statement}
Let $X^0 = \{x_1^0, \ldots, x_n^0\}$ and $Y^0 = \{y_1^0, \ldots, y_n^0\}$ be two sets of points (the superscript emphasizes that these points are obtained at the start time $t = 0$) in the unit square $Q$ \footnote{Our results are scale-invariant because all the theorems hold for squares of any size with proper scaling. Hence, a unit square environment is used throughout the paper.}, selected uniformly randomly. Place $n = |X^0| = |Y^0|$ point robots on the points in $X^0$, with robot $a_i$ occupying $x_i^0$. Each robot has a unique integer label (e.g., $i$). In general, we denote robot $a_i$'s location (coordinates) at time $t \ge 0$ as $x_i(t)$. The basic task (to be formally defined) is to move the robots so that at some {\em final time} $t^f \ge 0$, every $y \in Y^0$ is occupied by a robot. We may assume that there is a final time $t_i^f$ for each robot $a_i$, such that $x_i(t) \equiv x_i(t_i^f)$ for $t \ge t_i^f$. For convenience, we also refer to $X^0$ and $Y^0$ as the set of initial locations and the set of target locations, respectively. 

\subsubsection*{Motion model} The robots are single integrators, {\em i.e.}, $\dot x_i(t) = u_i(t)$ with $u_i(t)$ being piece-wise smooth and $\Vert u_i(t) \Vert_2 \in \{0, 1\}$. We assume the size of the robots is negligible with respect to the distance they travel and ignore collisions between robots. 

\customlabel{cm:1}{1}
\customlabel{cm:2}{2}
\subsubsection*{Communication Model \ref{cm:1}} 
\begin{figure}[htp]
\begin{center}
\begin{tabular}{ccc}
    \includegraphics[width=1.5in]{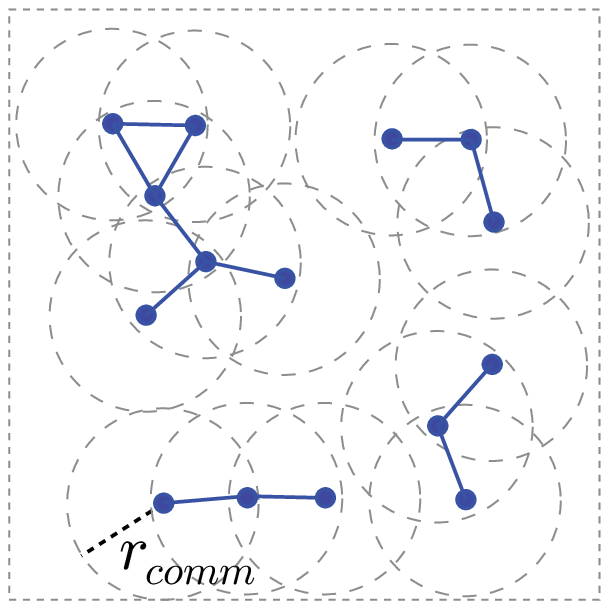} &&
    \includegraphics[width=1.5in]{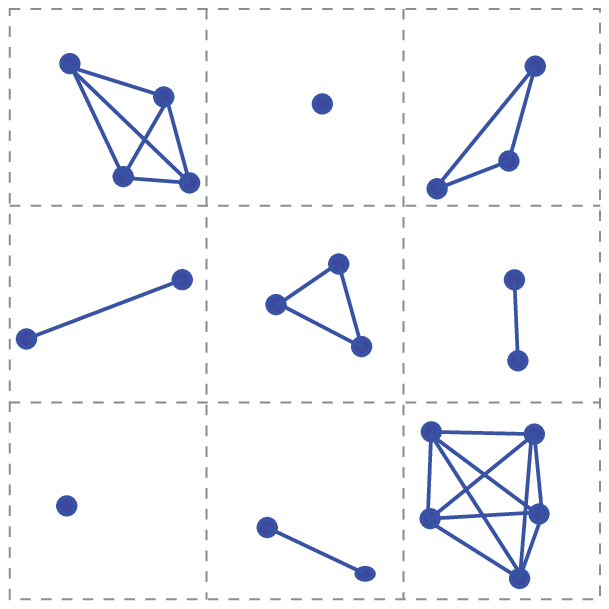} 
    \\ (a) Comm. model \ref{cm:1} &  &(b) Comm. model \ref{cm:2} \\ 
\end{tabular}
\end{center}
\caption{\label{fig:model} (a) The communication graph (solid blue nodes and edges) for a set of robots under Communication Model \ref{cm:1} with a communication radius of $r_{\mathrm{comm}}$. Robots (blue dots) in the same connected component of a communication graph can freely communicate with each other. (b) The communication graph for a set of robots under Communication Model \ref{cm:2} with $m = b^2 = 9$.}
\end{figure}

We study two communication models in this paper. In the first communication model, a robot $a_i$ may communicate with other robots within a disc of radius $r_{\mathrm{comm}}$ centered at $x_i(t)$. At any given time $t \ge 0$, we define the (undirected) {\em communication graph} $G(t)$, which is a geometric graph that changes over time, as follows. $G(t)$ has $n$ vertices $v_1, \ldots, v_n$, corresponding to robots $a_1, \ldots, a_n$, respectively. There is an edge between two vertices $v_i$ and $v_j$ if the corresponding robot locations $x_i(t)$ and $x_j(t)$, respectively, satisfy $\Vert x_i(t) - x_j(t) \Vert_2 \le r_{\mathrm{comm}}$. Figure \ref{fig:model}(a) provides an example of a (disconnected) communication graph. 

Given our focus on distance optimality, we make the simplifying assumption that all robots corresponding to vertices in a connected component of the communication graph may exchange information instantaneously. In other words, robots in a connected component of $G(t)$ can be treated as a single robot insofar as decision making is concerned. 

\subsubsection*{Communication Model \ref{cm:2}} The unit square $Q$ is divided into $m = b^2$ equal-sized smaller squares (regions).\footnote{In this paper, $m$ is frequently used to denote the number of small squares in a division of the unit square $Q$ and $b = \sqrt{m}$ is the number of resulting partitions on an edge of the unit square. The value of $m$ and $b$ may vary.} Robots within each region can communicate with one another but robots from different regions cannot exchange information (see, e.g., Fig. \ref{fig:model}(b)). This model mimics the natural (geometrical) resource allocation process in which supplies and demands are first matched locally; the surpluses and deficits within each region then get balanced out at larger regions, giving rise to a hierarchical strategy.  

\subsubsection*{Target-sensing model} We assume that a robot is aware of a point $y \in Y^0$ if $\Vert y - x_i(t) \Vert_2 \le r_{\mathrm{sense}}$, the {\em target-sensing radius}. 

The problem we consider in this paper is defined as follows. 

\begin{problem}[Target Assignment in Robotic Networks]\label{p:sa}Given $X^0$, $Y^0$, $r_{\mathrm{sense}}$, and Communication Model~\ref{cm:1} with $r_{\mathrm{comm}}$ or Communication Model~\ref{cm:2} , find a control strategy $\mathbf{u}(t) = [u_1(t), \ldots, u_n(t)]$, such that for some $0 \le t_i^f < \infty$ and some permutation $\sigma$ of the numbers $1, \ldots, n$, $x_i(t_i^f) = y_{\sigma(i)}^0$ for all $1 \le i \le n$. 
\end{problem}

Over all feasible solutions to an instance of Problem \ref{p:sa}, we are interested in minimizing the total distance traveled by all robots, which can be expressed as 
\begin{equation}\label{eq:sla}
D_n := \sum_{i = 1}^n \int_0^{t_i^f} \Vert\dot x_i(t)\Vert_2dt. 
\end{equation}

As an accurate proxy to the energy consumption of the entire system, the cost defined in (\ref{eq:sla}) is an appropriate objective in practice. Unless otherwise specified, {\em distance optimality} refers to minimizing $D_n$. Over all permutations $\sigma$ of the numbers $1, \ldots, n$, and for fixed $X^0$ and $Y^0$, the minimum total distance for robots moving along continuous paths is 
\begin{equation}\label{eq:dnstar}
D_n^* := \min_{\sigma} \sum_{i = 1}^n \Vert x_i^0 - y_{\sigma(i)}^0\Vert_2,
\end{equation}
which may or may not be achievable depending on the capabilities of the robots (e.g, if the robots cannot follow straight-line paths, then $D_n > D_n^*$). Let $\cal U$ denote the set of all possible control strategies that solve Problem \ref{p:sa} given a fixed set of capabilities for the robots, we say that distance optimality is achieved if  $\min_{\cal U} D_n = D_n^*$. Besides distance optimality, we also briefly discuss the total task completion time ({\em i.e.}, the sum of the individual task completion times as targets are occupied), denoted by $T_n$. If all robots start moving toward targets and do not stop in the middle, then $T_n = D_n$. In particular, we define $T_n^* := D_n^*$. 

\section{Guaranteeing Distance Optimality for Arbitrary $r_{\mathrm{comm}}$ and $r_{\mathrm{sense}}$}\label{sec:bound}
In this section, we use Communication Model \ref{cm:1}. In general, when $r_{\mathrm{sense}} < \sqrt{2}$ or $r_{\mathrm{comm}} < \sqrt{2}$, it is impossible to guarantee distance optimality, since global assignment is no longer possible in general. For example, as $r_{\mathrm{sense}} \to 0$, the robots must search for the targets before assignments can be made; it is very unlikely that the paths taken by the robots toward the targets will be straight lines, which is required to obtain $D_n^*$. This raises the following question: Given a pair of $r_{\mathrm{comm}}$ and $r_{\mathrm{sense}}$, under what conditions can we ensure distance optimality? Theorem \ref{t:sn} answers this question. 

\begin{theorem}\label{t:sn} In a unit square, under sensing and communication constraints ({\em i.e.}, $r_{\mathrm{comm}}, r_{\mathrm{sense}} < \sqrt{2}$), distance optimality can be achieved with probability one if and only if at $t = 0$: 
\begin{enumerate}
\item[{\em i)}] the communication graph is connected, and 
\item[{\em ii)}] every target is within a distance of $r_{\mathrm{sense}}$ to some robot. 
\end{enumerate}
\end{theorem}
{\sc Proof}. We first prove that the conditions are necessary with two claims: 1) an optimal assignment that minimizes $D_n$ is possible in general only if $G(0)$ is connected, and 2) an optimal assignment that minimizes $D_n$ is possible only if for all $y \in Y^0$, $y$ is within a distance of $r_{\mathrm{sense}}$ to some $x \in X^0$.

\begin{figure}[htp]
\begin{center}
    \includegraphics[width=2.25in]{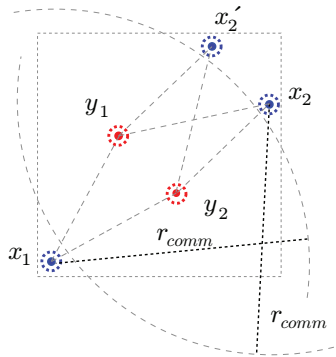} 
\end{center}
\caption{\label{fig:necc} A general setup in which the two robots cannot communicate with each other at $t = 0$ and therefore cannot always decide an optimal assignment at $t =0$.}
\end{figure}

To see that the first claim is true, we note that distance-optimal assignments forbid robots from moving unnecessarily, requiring at $t=0$ a pairing between elements of $X^0$ and $Y^0$ that minimizes $D_n$. We now show that this is not possible in general when $r_{\mathrm{comm}} < \sqrt{2}$. For $n = 2$, assume that the two targets are located at $y_1$ and $y_2$ as given in Fig. \ref{fig:necc} (solid red dots). Assume the first robot $a_1$ is located at $x_1$ (the blue solid dot at the lower left of Fig. \ref{fig:necc}) and $a_1$ is of 
equal distance to $y_1$ and $y_2$. Let the second robot $a_2$ take two possible locations $x_2$ and $x_2'$ as shown, which are symmetric along a diagonal of $Q$. If $a_2$ is at $x_2$ (resp. $x_2'$), then $a_2$ should go to $y_2$ (resp. $y_1$), forcing $a_1$ to go to $y_1$ (resp. $y_2$). Not knowing $a_2$'s location because $a_1$ is out of $a_2$'s communication radius, $a_1$ has at most $50\%$ chance of picking the distance minimizing choice at $t = 0$. We can readily extend the locations of the robots and targets to include neighborhoods around them (the dotted circles in Fig. \ref{fig:necc}) to show that there is a non-zero probability that an optimal assignment cannot be made at $t = 0$. This proves that that $G(0)$ cannot have more than one connected component and must be connected. The example can be extended to work for arbitrary $n$ by adding additional robots and targets to close vicinities of $x_1$ and $y_1$, respectively.  

For the second claim, suppose that at $t = 0$, some $y \in Y^0$ is not within a distance of $r_{\mathrm{sense}}$ to any $x \in X^0$. A robot must move to search for that $y$. This will cause the robot to follow a path that is not a straight line with probability one, implying that $D_n = D_n^*$ with probability zero.

It is not hard to see that the necessary conditions from the two claims are also sufficient: when $G(0)$ is connected and each target is observable by some robot $a_i$, the robots can decide at $t=0$ a global assignment that minimizes $D_n$. ~\qed

Theorem \ref{t:sn} suggests a simple way for ensuring distance optimality by either increasing the number of robots or increasing one or both of $r_{\mathrm{comm}}$ and $r_{\mathrm{sense}}$. This essentially leads to a centralized communication and control strategy (Strategy \ref{alg:cen}). Note that given the assignment permutation $\sigma$, each robot $a_i$ can easily compute its straight-line path between $x^0_i$ and $y^0_{\sigma(i)}$. Since every robot can carry out the computation in Strategy \ref{alg:cen}, to resolve conflicting decisions and avoid unnecessary computation, we may let the highest labeled robot (e.g., $a_n$) handle the entire assignment process. 

\def\cent{{\sc Centralized Assignment}}
\IncMargin{0.2em}
\begin{algorithm}\label{alg:cen}
\SetAlgoVlined
\KwIn{$X^0, Y^0$} 
\KwResult{permutation $\sigma$ that assigns a robot $a_i$ to $y^0_{\sigma(i)}$}
\BlankLine
 compute $d_{i,j} = \Vert x_i - y_j \Vert_2$ between each pair of $(x_i, y_j)$ in which $x_i \in X^0$ and $y_j \in Y^0$ \\
 compute over $\{d_{i,j}\}$ an assignment that minimizes $D_n$\\
 communicate the assignment to all the robots
 \caption{\cent}
\end{algorithm}
\DecMargin{0.2em}

The rest of this section establishes how the conditions from Theorem \ref{t:sn} can be met. We point out that similar conclusions can also be reached by exploring Theorem \ref{t:rgg}, which yields an asymptotic relationship between the required number of robots for $G(0)$ to be connected and $r_{\mathrm{comm}}$. We take a different approach and produce the required number of robots as an explicit function of $r_{\mathrm{comm}}$ without the asymptotic assumption. 

\subsection{Guaranteeing a Connected $G(0)$}
Since the robots can be anywhere in the unit square $Q$, given a communication radius of $r_{\mathrm{comm}} < \sqrt{2}$, intuitively, at least $\Theta(1/r_{\mathrm{comm}}^2)$ robots are needed for a connected $G(0)$, which requires the robots to take a lattice-like formation such as a grid. It turns out that when the robots are uniformly randomly distributed, only a logarithmic factor more robots are needed to ensure a connected $G(0)$. 

\begin{figure}[htp]
\begin{center}
    \includegraphics[width=2.251in]{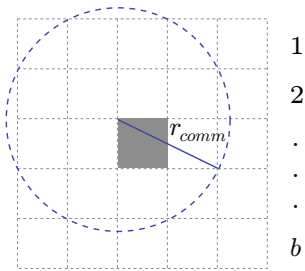}
\end{center}
\caption{\label{fig:upper} If the small squares have a side length of $\lceil \sqrt{5}/r_{\mathrm{comm}} \rceil$ or smaller, then a robot in such a square (e.g., the gray square) can communicate with any robot in the four neighboring small squares sharing a side with the gray square.}
\end{figure}

\begin{lemma}\label{l:comm2}Suppose that $n$ robots are uniformly randomly distributed in the unit square. For fixed $r_{\mathrm{comm}} < \sqrt{2}$ and $0 < \epsilon < 1$,  at $t = 0$, the communication graph is connected with probability at least $1 - \epsilon$ if 
\begin{equation}\label{eq:comm2}
n \ge \lceil\frac{\sqrt{5}}{r_{\mathrm{comm}}}\rceil^2\log (\frac{1}{\epsilon}\lceil\frac{\sqrt{5}}{r_{\mathrm{comm}}}\rceil^2). 
\end{equation}
\end{lemma}
\noindent {\sc Proof.} We divide the unit square $Q$ into $m = b^2$ equal-sized small squares with $b = \lceil \sqrt{5}/r_{\mathrm{comm}} \rceil$. Label these small squares $\{q_1, \ldots, q_{m}\}$. Under this division scheme, a robot residing in a small square $q_i$ can communicate with any other robot in the four squares sharing a side with $q_i$ (see Fig. \ref{fig:upper}). Therefore, $G(0)$ is connected if each $q_i$ contains a robot. Let $n_i$ denote the number of robots in $q_i$. Then
\begin{displaymath}
\mathbf{P}(n_i = 0) = (1 - \frac{1}{m})^n < e^{-\frac{n}{m}}.
\end{displaymath}

The inequality holds because $(1 - x)^n < e^{-nx}$ for $0 < x < 1$. To see this, let $f(x) = \log (1-x)/x$. The Taylor expansion of $f(x)$ at $x=0$ 
is $-1 - x/2 - x^2/3 + o(x^3) < -1$ for $0 < x < 1$. This shows that $\log(1-x) < -x$ for $0 < x < 1 \Rightarrow n\log(1-x) < -nx \Rightarrow (1-x)^n < e^{-nx}$. By Boole's inequality ({\em i.e.}, the union bound), the probability that at least one of $q_1, \ldots, q_m$ is empty can be upper bounded as
\begin{displaymath}
\mathbf{P}(\bigcup_{i=1}^m E(n_i = 0)) \le \sum_{i = 1}^m\mathbf{P}(n_i = 0) < me^{-\frac{n}{m}}.
\end{displaymath}

Setting $me^{-{n}/{m}} = \epsilon$ and replacing $m = \lceil \sqrt{5}/r_{\mathrm{comm}} \rceil^2$ yields 
\begin{displaymath}
\begin{array}{l}
\displaystyle\lceil\frac{\sqrt{5}}{r_{\mathrm{comm}}}\rceil^2\textrm{exp}(-n\frac{1}{\lceil\frac{\sqrt{5}}{r_{\mathrm{comm}}}\rceil^2}) = \epsilon \\ 
\displaystyle\Rightarrow n = (\lceil\frac{\sqrt{5}}{r_{\mathrm{comm}}}\rceil^2)\log (\frac{1}{\epsilon}\lceil\frac{\sqrt{5}}{r_{\mathrm{comm}}}\rceil^2), 
\end{array}
\end{displaymath} 
which guarantees that each small square contains at least one robot with probability $1 - \epsilon$. ~\qed

The bound in Lemma \ref{l:comm2} can be further tightened; Corollary~\ref{c:tighter} (below) illustrates one way to achieve this. It produces $n$ smaller than that given by \eqref{eq:comm2} when $r_{\mathrm{comm}} < \sqrt{5}/2$. 

\begin{corollary}\label{c:tighter}Suppose that $n$ robots are uniformly randomly distributed in the unit square. For fixed $r_{\mathrm{comm}} < \sqrt{2}$ and $0 < \epsilon < 1$, at $t = 0$, the communication graph is connected with probability at least $1 - \epsilon$ if
\begin{equation}\label{eq:comm3}
n \ge \lceil\frac{\sqrt{5}}{r_{\mathrm{comm}}}\rceil^2\log \Big[\frac{1}{\epsilon}(\frac{1}{2}\lceil\frac{\sqrt{5}}{r_{\mathrm{comm}}}\rceil^2 + \lceil\frac{\sqrt{5}}{r_{\mathrm{comm}}}\rceil)\Big].
\end{equation}
\end{corollary}
\begin{figure}[htp]
\begin{center}
    \includegraphics[width=2.251in]{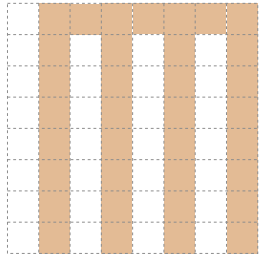}
\end{center}
\caption{\label{fig:uf} As long as each of the shaded small squares contains an robot, $G(0)$ must be connected. Therefore, only $b^2/2 + b$ small squares need to have robots in them.}
\end{figure}

\noindent {\sc Proof.} If each of the shaded small squares in Fig. \ref{fig:uf} has at least one robot, then $G(0)$ must be connected: any robot falling in a small white square must be connected to some robot in a shaded small square. This shows that (\ref{eq:comm3}) is sufficient.~\qed 

{\bf Remark.} In comparison to Theorem \ref{t:rgg}, Lemma \ref{l:comm2} provides $n$ as an explicit function of $r_{\mathrm{comm}}$. Moreover, our sufficient condition on $n$ given in (\ref{eq:comm2}) (and (\ref{eq:comm3})), unlike (\ref{eq:rgg}), is not an asymptotic bound. Therefore, our bound applies to an arbitrary $r_{\textrm{comm}}$. On the other hand, if we let $r_{\mathrm{comm}} \to 0$, then an asymptotic statement can also be made. 

\begin{lemma}\label{l:comm}Suppose that $n$ robots, each with a communication radius of $r_{\mathrm{comm}}$, are uniformly randomly distributed in the unit square. At $t = 0$, the communication graph is asymptotically connected with arbitrarily high probability $e^{-e^{-c}}$ (for some $c > 0$) if \begin{equation}\label{eq:comm}
n \ge (2\log \lceil\frac{\sqrt{5}}{r_{\mathrm{comm}}}\rceil + c) \lceil\frac{\sqrt{5}}{r_{\mathrm{comm}}}\rceil^2. 
\end{equation}
\end{lemma}
\noindent {\sc Proof.} Given the division scheme used in the proof of Lemma \ref{l:comm2}, distributing robots into the unit square $Q$ is equivalent to tossing the robots (balls) into the $m$ small squares (bins) uniformly randomly. By Corollary \ref{c:er1}, as $m \to \infty$, having $n \ge m\log m + cm = (2\log \lceil\sqrt{5}/{r_{\mathrm{comm}}}\rceil + c) \lceil \sqrt{5}/{r_{\mathrm{comm}}}\rceil^2$ robots guarantees that all $m$ small squares must have at least one robot each with probability $e^{-e^{-c}}$.~\qed

Since $f(x) = cx$ grows slower than $g(x) = x\log x$ as $x \to \infty$, Lemma \ref{l:comm} says that $n = \Theta((1/{r_{\mathrm{comm}}})^2$ $\log(1/{r_{\mathrm{comm}}}))$ robots can ensure that $G(0)$ is connected with probability arbitrarily close to one asymptotically. Next, we show that these many robots are also necessary for the high probability guarantee. 

Let $\mathbf{P}_{n, m}(E)$ denote the probability of an event $E$ happening after tossing $n$ balls into $m$ bins. We work with two events: $E_0$, the event that ``at least one bin is empty'', and $E_1$, the event that ``at least one bin contains exactly one ball''. We want to show that $\mathbf{P}_{n, m}(E_1)$ is not too small for $n$ up to $m\log m$, which is proven in the next two lemmas. 

\begin{lemma}\label{l:er3} Suppose that $1 \le n \le m$ balls are tossed uniformly randomly into $m$ bins. Then
\begin{displaymath}
\mathbf{P}_{n, m}(E_1) \ge (1-\frac{1}{m})^{m-1} > e^{-1}.
\end{displaymath}
\end{lemma}
\noindent {\sc Proof.} First we prove a useful inequality: for $m \in \mathbb N$,
\begin{equation}\label{eminus1}
(1-\frac{1}{m})^{m-1} > e^{-1}.
\end{equation}
To see this, note that the function $\log(1-x)^{\frac{1}{x} - 1}$ has a Taylor expansion of $-1 + x/2 + o(x^2) > -1$ for small $x > 0$, yielding that $(1-x)^{\frac{1}{x} - 1} > e^{-1}$ for small $x > 0$. Since the derivative of $(1-x)^{\frac{1}{x} - 1}$ is positive for $x \in (0, 1)$, \eqref{eminus1} holds for all $m > 0$ (we use the definition $0^0 = 1$ here). 

To prove Lemma~\ref{l:er3}, because all bins are initially empty, after tossing the first ball, some bin contains exactly one ball. That is, $\mathbf{P}_{1, m}(E_1) = 1$. Let the bin occupied by the first ball be bin $1$. As $k - 1$ additional balls are tossed into the $m$ bins, the probability that none of these $k - 1$ balls occupy bin $1$ is $(1 - 1/m)^{k-1}$. Therefore, for $1 \le k \le m$, we have 
\begin{displaymath}
\begin{array}{l}
\mathbf{P}_{k, m}(E_1) \ge \displaystyle\mathbf{P}_{1, m}(E_1) (1 - \frac{1}{m})^{k-1}  \\ 
\qquad \ge \displaystyle\mathbf{P}_{1, m}(E_1) (1 - \frac{1}{m})^{m-1} \\
\qquad \displaystyle = (1 - \frac{1}{m})^{m - 1} > e^{-1}.
\end{array}
\end{displaymath} ~\qed

\begin{lemma}\label{l:er4} Suppose that $m < n < m \log m$ balls are tossed uniformly randomly into $m$ bins. As $m \to \infty$, 
\begin{displaymath}
\mathbf{P}_{n, m}(E_1) \ge (1-e^{-e})(1-\frac{1}{m})^{m-1} > (1-e^{-e})e^{-1}.
\end{displaymath}
\end{lemma}
\noindent {\sc Proof.} Suppose that after an experiment of $n'$ tosses into $m$ bins, $E_0$ holds; {\em i.e.}, at least one bin is empty. Without loss of generality, we assume the empty bin is bin $1$. Now consider tossing an additional $k$ balls into the $m$ bins. The probability of exactly one of these $k$ balls falling in bin $1$ is 
\begin{displaymath}
\begin{array}{l}
\quad \mathbf{P}_{k, m}(\textrm{exactly one ball falls in bin $1$}) \\
= \displaystyle\binom{k}{1}\frac{1}{m}(1-\frac{1}{m})^{k-1} = \frac{k}{m}(1-\frac{1}{m})^{k-1}. 
\end{array}
\end{displaymath}

Therefore, 
\begin{equation}\label{eq:er41}
\begin{array}{l}
\quad \displaystyle\mathbf{P}_{n'+k, m}(E_1) \\
\ge \displaystyle\mathbf{P}_{n', m}(E_0)\mathbf{P}_{k, m}(\textrm{exactly one ball falls in bin $1$}) \\
= \displaystyle\frac{k}{m}(1-\frac{1}{m})^{k-1}\mathbf{P}_{n', m}(E_0). 
\end{array}
\end{equation}

Letting $c = -1$ in Corollary \ref{c:er1}, we have
\begin{equation}
\lim_{m\to\infty}\mathbf{P}(T_1 \ge m \log m - m) = 1 - e^{-e}. 
\end{equation}

That is, as $m \to \infty$, for $0 < n' < m \log m - m$, $\mathbf{P}_{n', m}(E_0) \ge 1 - e^{-e}$. Plugging this into (\ref{eq:er41}) and letting $k = m$, we have that for $m < n < m\log m$, as $m \to \infty$,
\begin{displaymath}
\begin{array}{l}
\mathbf{P}_{n, m}(E_1) \ge \displaystyle (1 - e^{-e})\frac{m}{m}(1-\frac{1}{m})^{m-1} > (1 - e^{-e})e^{-1},
\end{array}
\end{displaymath}
in which the last inequality is by \eqref{eminus1}.
~\qed

Under the assumptions of Lemmas~\ref{l:er3} and~\ref{l:er4}, we always have that as $m \to \infty, \mathbf{P}_{n, m}(E_1) > \min\{e^{-1}, (1-e^{-e})e^{-1}\} > 0.34$. We now show that $n = \Theta((1/{r_{\mathrm{comm}}})^2\log(1/{r_{\mathrm{comm}}}))$ is a tight bound on the number of robots for guaranteeing the connectivity of $G(0)$ with high probability. 

\begin{theorem}\label{t:comm} For $n$ uniformly randomly distributed robots in a unit square with a communication radius $r_{\mathrm{comm}}$, 
\begin{equation}\label{eq:thetabound}
n = \Theta(\frac{1}{r_{\mathrm{comm}}^2} \log \frac{1}{r_{\mathrm{comm}}})
\end{equation} 
is necessary and sufficient to ensure that at $t =0$, the communication graph is asymptotically connected with arbitrarily high probability. 
\end{theorem}
\noindent {\sc Proof.} Lemma \ref{l:comm} covers sufficiency; we are to show that there is some non-trivial probability that $G(0)$ is disconnected if the number of robots satisfies 
\begin{displaymath}
n = o(\frac{1}{r_{\mathrm{comm}}^2} \log \frac{1}{r_{\mathrm{comm}}}).
\end{displaymath}

\begin{figure}[htp]
\begin{center}
    \includegraphics[width=2.25in]{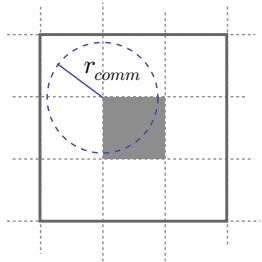}
\end{center}
\caption{\label{fig:lower} A $3 \times 3$ block as defined in the proof Theorem \ref{t:comm}.}
\end{figure}

To prove the claim, we partition the unit square $Q$ into $m = b^2$ equal-sized small squares in which $b = \lfloor 1.1/r_{\mathrm{comm}} \rfloor$. The factor of $1.1$ in the expression makes the side of the small square larger than $r_{\mathrm{comm}}$. Assuming that $m$ is divisible by $3$ (it is always possible to truncate some small squares to satisfy this), we may group the small squares into $m/9$ groups of $3 \times 3$ blocks (see, e.g., Fig. \ref{fig:lower}). 

If there is a single robot in a $3\times 3$ block, the robot cannot communicate with the rest of the robots if it falls inside the small square in the center of the block (e.g., the solid gray square in Fig. \ref{fig:lower}). By Lemmas \ref{l:er3} and \ref{l:er4}, for less than $(m/9)\log (m/9) = 2\lfloor{1.1}/{r_{\mathrm{comm}}}\rfloor^2 \log (\lfloor{1.1}/{r_{\mathrm{comm}}}\rfloor/3)/9$ robots, the probability of having at least one of these $3\times 3$ blocks containing exactly one robot is at least 0.34 as $m \to \infty$ ({\em i.e.}, $r_{\mathrm{comm}} \to 0$). If a $3\times 3$ block has exactly one robot in it, with probability of $1/9$, the robot is in the middle square. Therefore, with probability at least $0.34/9 \approx 0.04$, $G(0)$ is disconnected. 
~\qed

\subsection{Ensuring Target Observability}

With a connected communication graph $G(0)$ guaranteed by Lemma \ref{l:comm2}, we can solve a single assignment problem if for each $y \in Y^0$, $\Vert y - x \Vert_2 \le r_{\mathrm{sense}}$ for some $x \in X^0$. Similar techniques used in the proof of Lemma \ref{l:comm2} lead to a similar lower bound on $n$. 

\begin{lemma}\label{l:sense2} Suppose that $n$ robots and $n$ targets are uniformly randomly distributed in the unit square. For fixed $r_{\mathrm{sense}}$ and $0 < \epsilon < 1$, every target is observable by some robot at $t = 0$ with probability at least $1 - \epsilon$ if 
\begin{equation}\label{eq:t3}
n \ge \lceil\frac{\sqrt{2}}{r_{\mathrm{sense}}}\rceil^2\log (\frac{1}{\epsilon}\lceil\frac{\sqrt{2}}{r_{\mathrm{sense}}}\rceil^2).
\end{equation}
\end{lemma}
\noindent {\sc Proof.} If we partition the unit square $Q$ into $\lceil \sqrt{2}/r_{\mathrm{sense}} \rceil^2$ equal-sized small squares and there is at least one robot in each small square, then any point of $Q$ is within $r_{\mathrm{sense}}$ distance to some robot. Following the same argument used in the proof of Lemma \ref{l:comm2}, the inequality from (\ref{eq:t3}) ensures that this happens with probability at least $1 - \epsilon$. ~\qed

Putting together Lemmas \ref{l:comm2} and \ref{l:sense2}, we obtain a lower bound on $n$ that makes a distance-optimal assignment possible. 
\begin{theorem}\label{t:ob} Suppose that $n$ robots and $n$ targets are uniformly randomly distributed in the unit square. Fixing $0 < \epsilon < 1$, at $t = 0$, the communication graph is connected and every target is observable by some robot with probability at least $1 - \epsilon$ if 
\begin{equation}\label{eq:sum}
n \ge \lceil \frac{\sqrt{10}}{\theta} \rceil^2\log (\frac{1}{\epsilon}\lceil \frac{\sqrt{10}}{\theta} \rceil^2),
\end{equation}
in which $\theta := \min \{ \sqrt{5} r_{\mathrm{sense}}, \sqrt{2} r_{\mathrm{comm}} \}$.
\end{theorem}
\noindent {\sc Proof.} When $\theta = \sqrt{5} r_{\mathrm{sense}}$, \eqref{eq:sum} becomes (\ref{eq:t3}), which implies (\ref{eq:comm}). Therefore, $G(0)$ is connected with probability $1 - \epsilon$. 

When $\theta = \sqrt{2} r_{\mathrm{comm}}$, {\em i.e.}, $r_{\mathrm{sense}} \ge \sqrt{10}r_{\mathrm{comm}}/5$, by Lemma \ref{l:comm2}, (\ref{eq:comm}) implies that $G(0)$ is connected with probability $1 - \epsilon$. Moreover,  there is at least one robot in each of the small squares with a side length of at most $r_{\mathrm{comm}}/\sqrt{5}$ (as specified in the proof of Lemma \ref{l:comm2}). Having $r_{\mathrm{sense}} \ge \sqrt{10}r_{\mathrm{comm}}/5$ guarantees that robots in a small square observes all targes within the same small square. Therefore, every $y \in Y^0$ is within a distance of $r_{\mathrm{sense}}$ to some $x \in X^0$. ~\qed

{\bf Remark.} Theorem \ref{t:ob} is not an asymptotic result and applies to all $r_{\mathrm{comm}}$ and $r_{\mathrm{sense}}$. If a high probability asymptotic result is desirable, Lemma \ref{l:sense2} can be readily turned into a version similar to Theorem \ref{t:comm}, by following the same proof techniques. In view of this fact, the bounds from Theorem \ref{t:ob} are asymptotically tight. 

\section{Hierarchical Strategies for $r_{\mathrm{sense}} \ge \sqrt{2}$: Optimal Distance and Performance Guarantees}\label{sec:hier}

\def\hier{{\sc Hierarchical-Divide-and-Conquer}}
In this section, we work with the (region-based) Communication Model \ref{cm:2} and assume that $r_{\mathrm{sense}} \ge \sqrt{2}$ (that is, every robot is aware of the entire $Y^0$). The study of Communication Model \ref{cm:2}, besides leading to interesting conclusions on hierarchical strategies, also facilitates the analysis in Section \ref{sec:strat} as we revisit Communication Model \ref{cm:1}. 

A region-based communication model naturally leads to a hierarchical strategy for solving Problem~\ref{p:sa} under the optimality criterion of minimizing the cost defined by~\eqref{eq:sla}. Let $h \ge 1$ be the number of hierarchies and $m_i, 1 \le i \le h$, be the number of equal-sized regions at hierarchy $i$. We make the following assumptions that are mainly used in Theorem \ref{t:hier}: {\em i)} $m_1 \equiv 1$, {\em ii)} $m_{i + 1} \ge m_i$, and {\em iii)} a region at a higher numbered hierarchy is contained in a single region at a lower numbered hierarchy. For example, dividing $Q$ into $4^{i-1}$ squares at hierarchy $i$ satisfies these requirements. 
\IncMargin{0.2em}
\begin{algorithm}\label{alg:hier}
 \SetAlgoVlined
 \KwIn{$X^0, Y^0, h, m_1, \ldots, m_h$} 
 \KwResult{permutation $\sigma$ that assigns a robot $a_i$ to $y^0_{\sigma(i)}$}
\BlankLine
 \For{each hierarchy $i \in \{1, \ldots, h\}$ in decreasing order}{
 		\For{each region $j \in \{1, \ldots, m_i\}$}{
			  let $n_a$ and $n_g$ be the number of unmatched robots and targets in region $j$, respectively\\
			  \uIf{$n_a \ge n_g > 0$}{pick the first $n_g$ robots from the $n_a$ unmatched robots and
			  run an assignment algorithm to match them with the $n_g$ unmatched targets in region $j$
			  }
			  \uElseIf{$n_g \ge n_a > 0$}
			  {pick the first $n_a$ targets from the $n_g$ unmatched targets and
			  run an assignment algorithm to match the $n_a$ unmatched robots with these $n_a$ targets in region $j$}
			  \Else{continue}
		}
 }
 \caption{\hier}
\end{algorithm}
\DecMargin{0.2em}
We call the associated strategy under these assumptions the {\em hierarchical  divide-and-conquer} strategy, the details of which are described in Strategy \ref{alg:hier}. Note that for each region in Strategy \ref{alg:hier}, the robots can again let the highest labeled robot within the region carry out the strategy locally.

It is clear that Strategy \ref{alg:hier} is correct by construction because $|X^0| = |Y^0|$. The rest of this section is devoted to analyzing the strategy. We begin with a single hierarchy ($h = 1$). Since $r_{\mathrm{sense}} \ge \sqrt{2}$ implies that all robots are aware of the entire set $Y^0$, the robots may form a consensus of which robot should go to which target at $t = 0$ by finding an optimal assignment $\sigma$ that yields $D_n^*$ as defined by~\eqref{eq:dnstar}. This assignment problem can be solved using a bipartite matching algorithm such as the Hungarian method. Ajtai, Koml\'{o}s, and Tusn\'{a}dy proved the following about $D_n^*$.

\begin{theorem}[Optimal Matching \cite{AjtKomTus84}]\label{t:bound} Assuming that $n$ points are i.i.d. following the uniform distribution over a unit square, then, with probability $1 - o(1)$, 
\begin{equation}\label{eq:bound}
C_1\sqrt{n\log n} \le D_n^* \le C_2\sqrt{n\log n},
\end{equation}
in which $C_1$ and $C_2$ are positive constants. 
\end{theorem}
\begin{figure}[htp]
\begin{center}
    \includegraphics[width=3in]{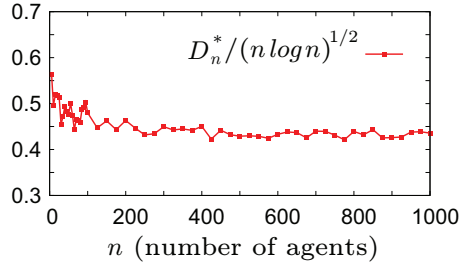}
\end{center}
\caption{\label{fig:nlogn} The ratio of $D_n^*/\sqrt{n\log n}$. Each data point is an average of 25 runs.}
\end{figure}

{\bf Remark.} The second inequality in~\eqref{eq:bound} remains true in expectation and also for arbitrary probability measures on $[0,1]^2$, albeit with a different universal constant than $C_2$, by a result of Talagrand \cite{Tal92}. Therefore, $D_n^* = \Theta(\sqrt{n \log n})$ in expectation. Although no formulas for $C_1$ and $C_2$ from \eqref{eq:bound} were given in \cite{AjtKomTus84}, a simulation study suggests that $C_1 < C_2 < 1$ and $C_2/C_1 \to 1$ as $n \to \infty$. As an example, for $200 \le n \le 10000$, $0.4\sqrt{n\log n} \le D_n^* \le 0.5 \sqrt{n\log n}$ on average (see Fig. \ref{fig:nlogn}). 

Next, we look at the general case with $h > 1$ hierarchies. To bound $D_n$, at each hierarchy $i$, we need to know the number of robots that cannot be matched locally. We derive this number in Lemma \ref{l:ran}. Note that Lemma \ref{l:ran} does not depend on $m$ and $n$ being large. 

\begin{lemma}\label{l:ran}Suppose that $n$ robots and $n$ targets are uniformly randomly distributed in the unit square $Q$, and $Q$ is divided into $m$ equal-sized regions. Within each of these $m$ regions, the robots are matched one-to-one with the targets until no more matchings can be made. The total number of robots that are left unmatched is no more than $\sqrt{n(m-1)/2}$ in expectation. 
\end{lemma}
\noindent {\sc Proof.} Restricting to one of the $m$ equal-sized regions, say $q_i$, we know for $x_j^0 \in X^0$ and $y_j^0 \in Y^0$, 
$$
\mathbf{P}(x_j^0 \in q_i) = \mathbf{P}(y_j^0 \in q_i) = \frac{1}{m},
$$ 
and 
$$
\mathbf{P}(x_j^0 \in q_i, y_j^0 \notin q_i) = \mathbf{P}(x_j^0 \notin q_i, y_j^0 \in q_i) = \frac{m-1}{m^2},
$$
in which the event $(x_j^0 \in q_i, y_j^0 \notin q_i)$ represents a surplus of one robot in $q_i$ and the event $(x_j^0 \notin q_i, y_j^0 \in q_i)$ a deficit in $q_i$. Thus, we may view the experiment of picking $x_j^0$ and $y_j^0$ as a one step walk on the real line starting at the origin, with $(m-1)/m^2$ probability of moving $\pm 1$. The entire process of picking $X^0$ and $Y^0$ can then be treated as a random walk of $n$ such steps. 

Under this random walk analogy, we may use a random variable $Z_j \in \{0, \pm 1\}$ to represent the outcome of picking $(x_j^0, y_j^0)$. We immediately have that $\mathbf{E}[Z_j^2] = 2(m-1)/m^2$. Letting $S_n = Z_1 + \ldots + Z_n$, we can compute the variance of $S_n$ as 
\begin{displaymath}
\begin{array}{l}
\mathbf{E}[S_n^2] = \displaystyle \mathbf{E}[(Z_1+\ldots+Z_n)^2] =  \displaystyle \mathbf{E}[Z_1^2 + \ldots + Z_n^2] \\
\quad =  \displaystyle n\mathbf{E}[Z_j^2] = \frac{2n(m-1)}{m^2}.
\end{array}
\end{displaymath}

Applying Jensen's inequality to the concave function $\sqrt{x}$ with $x = |S_n|^2 = S_n^2$, we have 
$$
\begin{array}{l}
\displaystyle \mathbf{E}[|S_n|]  = \mathbf{E}[\sqrt{S_n^2}] \le \sqrt{\mathbf{E}[S_n^2]}\\
\displaystyle\qquad\Rightarrow \mathbf{E}[|S_n|] \le \sqrt{\frac{2n(m-1)}{m^2}}.
\end{array}
$$

Because, in expectation, an equal number of the $m$ regions have surpluses (more robots than targets) and deficits (fewer robots than targets), and some of the $m$ regions may have neither, no more than half of the $m$ regions should have a surplus of robots on average. The total number of unmatched robots in expectation is then no more than $(m/2)*\mathbf{E}[|S_n|] \le \sqrt{n(m-1)/2}.$ ~\qed

The distance traveled by the matched robots at the bottom hierarchy with $m$ regions can be bounded easily. For simplicity, we now assume that these regions are equal-sized squares.  
\begin{lemma}\label{l:jensen}Suppose that $n$ robots and $n$ targets are uniformly randomly distributed in the unit square $Q$, and $Q$ is divided into $m$ equal-sized small squares. Within each of these $m$ small squares, the robots are matched one-to-one with the targets until no more matchings can be made. The minimum total distance of matchings made between the robots and the targets within the small squares is no more than $C\sqrt{n\log n}$ in expectation, for some positive constant $C$. 
\end{lemma}
\noindent {\sc Proof.} Since $Q$ is divided into $m$ squares, these squares all have a side length of $1/\sqrt{m}$. Let one such square be $q_i$ with $n_i$ robots (note that $\sum_{i=1}^m n_i = n$). Since a uniform distribution restricted to $q_i$ is again uniform, we can apply Theorem~\ref{t:bound} to $q_i$.  If we let these $n_i$ robots match only with targets inside $q_i$, then the total distance incurred locally will not exceed $C\sqrt{n_i\log n_i/m}$ in expectation. Here $C$ is some positive constant. 

Note that it is not necessarily the case that all $n_i$ robots will be matched locally in $q_i$. This does not affect the current proof. For some $1 \le i \le m$, it may be the case that no local matchings can be made because either $n_i = 0$ or there is no target in $q_i$. Let $m' \le m$ denote the number of these $m$ squares in which local matchings can be made. The total distance incurred by local matchings is then upper bounded by (note that $n_i$ is now indexed with respect to these $m'$ squares) 
$$
\displaystyle\sum_{i = 1}^{m'} C\sqrt{\frac{n_{i} \log n_{i}}{m}} = C\frac{m'}{\sqrt{m}}\sum_{i = 1}^{m'} \frac{1}{m'}\sqrt{n_{i} \log n_{i}}.
$$

Here we assume that $m' > 0$, otherwise the local matchings would have a distance cost of zero. Since the function $\varphi(x) = \sqrt{x \log x}$ is concave, by Jensen's inequality, $\mathbf{E}[\sqrt{x \log x}] $ $\le \sqrt{\mathbf{E}[x] \log (\mathbf{E}[x])}$. Letting $x = n_i$ and the expectation be carried out over the discrete uniform distribution with $1/m'$ probability each, we have
$$
\begin{array}{l}
\displaystyle C\frac{m'}{\sqrt{m}}\sum_{i = 1}^{m'} \frac{1}{m'}\sqrt{n_{i} \log n_{i}} \le C\frac{m'}{\sqrt{m}}\sqrt{(\sum_{i=1}^{m'}\frac{n_{i}}{m'}) \log (\sum_{i=1}^{m'}\frac{n_{i}}{m'})} \\
\displaystyle \qquad = C\sqrt{\frac{m'}{m}}\sqrt{(\sum_{i=1}^{m'}n_i) (\log (\sum_{i=1}^{m'} n_i) - \log(m'))} \\
\qquad \displaystyle\le C\sqrt{n \log n}.
\end{array}
$$ ~\qed

{\bf Remark.} With minor modifications, Lemma \ref{l:jensen} can be applied to regions with shapes other than squares. Defining the diameter of a two-dimensional region as the diameter of the region's smallest enclosing circle, the main requirement for the modification to work is that the maximum diameter of these regions is $O(1/\sqrt{m})$. 

We now give an upper bound on $D_n$, in expectation, for general hierarchical strategies. 

\begin{theorem}\label{t:hier} Suppose that $n$ robots and $n$ targets are uniformly randomly distributed in the unit square $Q$, and $Q$ is divided into $m_i$ equal-sized small squares at hierarchy $i$ with a total of $h \ge 2$ hierarchies. For all applicable $i \ge 1$, assume that $m_{i+1} \ge m_i$ and any small square at hierarchy $i + 1$ falls within a single square at hierarchy $i$. Then Strategy~\ref{alg:hier} yields
\begin{equation}\label{eq:dch}
\mathbf{E}[D_n] \le C\sqrt{n\log n} + \sum_{i = 1}^{h-1} \sqrt{\frac{nm_{i+1}}{m_i}}. 
\end{equation}
\end{theorem}
\noindent {\sc Proof.} The $C\sqrt{n\log n}$ term on the RHS of (\ref{eq:dch}) is due to Lemma \ref{l:jensen}. Then at each hierarchy $i$ with $1 \le i < h$, each of the $m_i$ squares contains $m_{i+1}/m_i$ smaller squares from hierarchy $i + 1$. Here we use the assumption that a region at a higher numbered  hierarchy falls completely within a single region at a lower numbered hierarchy. This means that a robot that gets matched at hierarchy $i$ needs to travel at most a distance of $\sqrt{2/m_i}$. Since there are no more than $\sqrt{n(m_{i+1}-1)/2} < \sqrt{m_{i+1}n/2}$ unmatched robots at hierarchy $i$ in expectation by Lemma~\ref{l:ran}, the distance incurred at hierarchy $i$ is no more than $\sqrt{nm_{i+1}/m_i}$ for $1 \le i < h$. Summing up all the distances then gives us the inequality~\eqref{eq:dch}. ~\qed

Theorem \ref{t:hier} allows us to upper bound the performances of different hierarchical strategies depending on the choices of $h$ and $\{m_i\}$. We observe that for fixed $h$ and $\{m_i\}$ independent of $n$, the first term $C \sqrt{n\log n}$ dominates the other terms in (\ref{eq:dch}) as $n \to \infty$. This implies that Strategy \ref{alg:hier} yields assignments whose total distance is at most a constant multiple of the optimal distance. This observation is summarized in Corollary~\ref{corollary:constant-optimality}. Recall that $D_n^*$ is the minimum possible distance defined by~\eqref{eq:dnstar}. 

\begin{corollary}\label{corollary:constant-optimality}For fixed $h$ and $m_1, \ldots, m_h$ that do not depend on $n$, as $n \to \infty$, Strategy \ref{alg:hier} yields target assignments with $D_n/D_n^* = O(1)$ in expectation. 
\end{corollary}

For example, with $h \ge 2$ and $m_i = 4^{i - 1}$ at hierarchy $i$, we have
\begin{equation}\label{eq:dch1}
\begin{array}{l}
\mathbf{E}[D_n] \le \displaystyle C\sqrt{n\log n} + \sum_{i = 1}^{h-1}\sqrt{4n}\\
\quad\,\,\, = \displaystyle C\sqrt{n\log n} + 2(h - 1)\sqrt{n}.
\end{array}
\end{equation}

For any fixed $h$, as $n \to \infty$, $D_n/D_n^* \le C/C_1 + o(1) = O(1)$. A constant approximation ratio can also be achieved when $h$ and $\{m_i\}$ depend on $n$. For example, letting $h = 3$, $m_2 = \log n$, and $m_3 = \log^2n$, we have
\begin{equation}\label{eq:dch2}
\mathbf{E}[D_n] \le \displaystyle C\sqrt{n\log n} + \sum_{i = 1}^{2}\sqrt{n\log n} =  (C + 2)\sqrt{n\log n}.
\end{equation}

Since hierarchical strategies need not run centralized assignment algorithms for all robots, the computational part of these strategies can be significantly faster. We will come back to this point in the next section. 

{\bf Remark.} Before concluding this section, it is worth mentioning that the results of this section continue to hold in only slightly weaker forms when the point sets $X^0, Y^0$ are drawn {\em i.i.d.} from the same {\em arbitrary distribution} over $[0,1]^2$ (based on Talagrand \cite{Tal92}). Since the topic of arbitrary probability measures diverges from the main focus of this paper, we only briefly discuss extending the results of this section to deal with arbitrary probability measures on $[0,1]^2$.  

To adapt Lemma~\ref{l:ran} for arbitrary probability measures, assume that each region $q_i$ (see the proof of Lemma~\ref{l:ran}) has an overall probability of $p_i$ of receiving a robot or target. Note that $\sum_{i=1}^m p_i = 1$. This changes the upper bound of $\mathbf{E}[|S_n|]$ for the region $q_i$ to $\sqrt{2np_i(1-p_i)}$. Then, over all $m$ regions, the total number of unmatched robots is bounded by 
\begin{displaymath}
\begin{array}{l}
\displaystyle\sum_{i=1}^m\sqrt{2np_i(1-p_i)} = m\sqrt{2n}\sum_{i=1}^m\frac{1}{m}\sqrt{p_i(1-p_i)}\\
\displaystyle\qquad \le m\sqrt{2n}\sqrt{\sum_{i=1}^m\frac{p_i}{m}(1-\sum_{i=1}^m\frac{p_i}{m})} \\
\displaystyle\qquad = m\sqrt{2n}\sqrt{\frac{1}{m}(1-\frac{1}{m})} \\
\displaystyle\qquad = \sqrt{2n(m-1)},
\end{array}
\end{displaymath}
in which the inequality is obtained by applying Jensen's inequality to the concave function $\sqrt{x(1-x)}$. 

Besides updating the uniform distribution of $X^0$ and $Y^0$ to an arbitrary probability measure, the statement and proof of Lemma \ref{l:jensen} remain largely unchanged. This is because the second inequality in \eqref{eq:bound} does not change asymptotically as the underlying robot and target distribution changes. Then, the inequality~\eqref{eq:dch} from Theorem~\ref{t:hier} merely adds a multiplicative constant of $2$ to its second term on the RHS. Because the first inequality in~\eqref{eq:bound} is not known to hold for arbitrary probability measures, we do not have a parallel of Corollary~\ref{corollary:constant-optimality} for  arbitrary probability measures. Nevertheless, these bounds for arbitrary probability measures suggest that the uniform distribution is among the {\em worst} distributions for Problem~\ref{p:sa} under the optimality constraint of minimizing~\eqref{eq:sla}. This is because the uniform distribution leads to an optimal assignment distance of $\Omega(\sqrt{n\log n})$, and an arbitrary distribution leads to an optimal assignment distance of $O(\sqrt{n\log n})$. Note that these updates also apply to the results in the next section with appropriate modifications. 

\section{Near Optimal Strategies}\label{sec:strat}

After exploring hierarchical strategies for the region-based Communication Model \ref{cm:2}, we now return to the range-based Communication Model \ref{cm:1}. If $r_{\mathrm{comm}}$ is arbitrary and the conditions specified in Theorem \ref{t:sn} are not known to hold, the best we can do is obtain near distance-optimal strategies. In this section, we show that constant ratio approximation of distant optimality is possible for arbitrary $r_{\mathrm{sense}}$ and $r_{\mathrm{comm}}$. The basic idea behind our strategies is to move the robots to pass around information about the locations of other robots. The assumption $r_{\mathrm{sense}} \ge \sqrt{2}$ is made temporarily. At the end of this section, we show how to remove this assumption without affecting asymptotic optimality. 

\subsection{Near Distance-Optimal Rendezvous Strategy}
\begin{figure}[htp]
\begin{center}
    \includegraphics[width=2.25in]{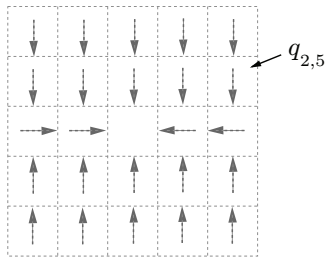}
\end{center}
\caption{\label{fig:cs} Directions for robots to move in the rendezvous strategy.}
\end{figure}

Our first suboptimal strategy uses moving robots for information aggregation until some robot is aware of the locations of all robots ({\em i.e.}, the set $X^0$), at which point a centralized optimal assignment can be made. Although some robots will move and change their locations during this process, the moved robots nevertheless are aware of their initial locations in $X^0$. To carry out the strategy, the unit square $Q$ is divided into $m = b^2$ disjoint, equal-sized small squares, with $b = \lceil \sqrt{2}/r_{\mathrm{comm}} \rceil$. These small squares are labeled as $q_{i,j}$'s, in which $i$ and $j$ are the row number and column number of the square, respectively (see Fig. \ref{fig:cs}).

Based on its initial location, each robot can identify the small square $q_{i, j}$ it lies in. At $t = 0$, the robots in the squares on row $1$ and row $b$ start moving in the direction as indicated in Fig. \ref{fig:cs}. We want to use these robot to pass the information of where all robots are. At most one robot per square is required to move since all robots in a small square can communicate with each other by the assumption $b = \lceil \sqrt{2}/r_{\mathrm{comm}} \rceil$. 

Assuming that a robot in a square $q_{i,j}$ is moving downwards, it keeps moving until it is within the communication radius of a robot in a square with label $q_{i+k,j}, k \ge 1$, at which point it passes over the information it has and stops. The robot in $q_{i+k,j}$ then does the same. The procedure continues until a robot reaches the middle of $Q$ (row $\lceil b/2 \rceil$). Then, the robots in the squares on row $\lceil b/2 \rceil$ repeat the same process horizontally until a robot in the center of $Q$ knows the locations of all other robots. At this point, the robot in the center of $Q$ that knows the location of all other robots makes a global assignment so that each robot is matched with a target. The moved robots then reverse their travel directions to deliver the assignment information to all robots. The outline of the strategy is given in Strategy \ref{alg:rend}.

\def\rend{{\sc Rendezvous}}
\IncMargin{0.2em}
\begin{algorithm}\label{alg:rend}
 \SetAlgoVlined
 \KwIn{$X^0, Y^0, r_{\mathrm{comm}}$} 
 \KwResult{produces a permutation $\sigma$ that assigns robots to targets and communicates $\sigma$ to all the robots}
\BlankLine
each robot computes its square $q_{i,j}$ based on $r_{\mathrm{comm}}$. Let the highest labeled robot within each $q_{i,j}$ be $a_{i,j}$, which represents $q_{i,j}$ \\
\For{each $q_{i,j}$, $1 \le i, j \le b = \lceil \sqrt{2}/r_{\mathrm{comm}} \rceil$}{
	  \uIf{$i \ne \lceil b/2 \rceil$} {$waitTime \leftarrow |\lceil b/2 \rceil - i|/b$}
		\Else{$waitTime \leftarrow1/2 + |\lceil b/2 \rceil - j|/b$}
				  $a_{i,j}$ waits for up to $waitTime$ units of time for information from a robot coming from the previous square. After the information is received or after $waitTime$ passes, $a_{i,j}$ starts moving to the next squares and delivers its information once it can communicate with another robot in these squares. It then stops\\
			  }
			  robot $a_{\lceil b/2 \rceil, \lceil b/2 \rceil}$ computes $\sigma$; the earlier communication process is then reversed to deliver $\sigma$ to all the robots. 
 \caption{\rend}
\end{algorithm}
\DecMargin{0.2em}

The correctness of Strategy~\ref{alg:rend} as an algorithm is proven by construction. Besides the distance cost from the assignment, the robots in each column travel at most a total distance of two. The middle row incurs an extra distance of at most two. Thus, in expectation, $D_n < D_n^* + 2b + 2$. Since $D_n^* = \Theta(\sqrt{n\log n})$, $D_n^*$ dominates $2b + 2$ when $b = o(\sqrt{n\log n})$. In particular, $n = \Theta(1/r_{\mathrm{comm}}^2)$ satisfies this requirement. Therefore, Strategy \ref{alg:rend} can yield near distance-optimal solution without requiring an $n$ as large as (\ref{eq:thetabound}) with respect to $1/r_{\mathrm{comm}}$. 

A drawback of Strategy \ref{alg:rend} is that no robot can move to the targets until the assignment phase is complete. This yields a total task completion time of $T_n \approx 2n + T_n^*$ in expectation, which is undesirable since $T_n^* = O(\sqrt{n\log n})$ asymptotically. Furthermore, Strategy \ref{alg:rend} requires running a centralized assignment algorithm for all robots. This might be impractical for large $n$. We address these issues with decentralized hierarchical strategies. 

\subsection{Decentralized Hierarchical Strategies}

We first look at a decentralized hierarchical strategy that combines Strategies \ref{alg:hier} and \ref{alg:rend}. Instead of waiting for a centralized assignment to be made, in each of the small square $q_{i,j}$ as specified in Strategy \ref{alg:rend}, we let the robots in $q_{i,j}$ be assigned to targets that belong to the same square (we refer to these as {\em local assignments}). The robots that are not matched to targets then carry out Strategy \ref{alg:rend}. We denote this hierarchical rendezvous strategy as Strategy \customlabel{algs4}{4}\ref{algs4} and omit the pseudo code. 

\begin{corollary}For Strategy \ref{algs4} (2-level Hierarchical Rendezvous), as $n \to \infty$,
\begin{equation}\label{eq:dch4}
\mathbf{E}[D_n] \le C\sqrt{n\log n} + \sqrt{{mn}} + 2\sqrt{m} + 2, 
\end{equation}
and
\begin{equation}\label{eq:tn1}
\mathbf{E}[T_n] = \displaystyle\Theta(\sqrt{n\log n} + \sqrt{mn}).
\end{equation}
\end{corollary}
\noindent {\sc Proof.} The bound on $\mathbf{E}[D_n]$, given by (\ref{eq:dch4}), is straightforward to compute using Theorem \ref{t:hier}, in which the first two terms on the right side of (\ref{eq:dch4}) correspond to the first and second terms of the right side of (\ref{eq:dch}), respectively, and the last two terms are due to communication overhead. For total completion time, all but $\Theta(\sqrt{mn})$ robots can start moving to their targets at $t = 0$. For the $\Theta(\sqrt{mn})$ robots, they need to wait no more than two units of time each before moving to their targets. This gives us (\ref{eq:tn1}). ~\qed

{\bf Remark.} Similar to Strategy \ref{alg:rend}, for any fixed $m$, in expectation, $D_n/D_n^* = O(1)$ as $n \to \infty$. Moreover, in contrast to Strategy \ref{alg:rend}, for any fixed $m$, $T_n/T_n^* = O(1)$ in expectation. Suppose that a centralized algorithm requires $t(n)$ running time. Using the same centralized algorithm, Strategy \ref{algs4} has a running time of $O(mt(n/m) + t(\sqrt{mn}))$. If $t(n) = O(n^3)$ as given by the Hungarian method, then Strategy \ref{algs4} has a running time of $O(n^3/m^2 + (mn)^{3/2})$. Taking $n = 10000, m = 10$, for example, we get a $1000$-time speedup. 

\begin{figure}[htp]
\begin{center}
    \includegraphics[width=1.71in]{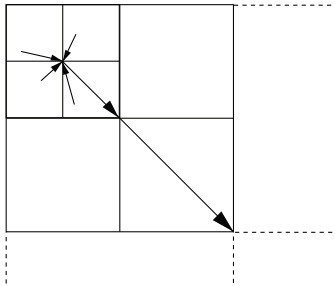} 
\end{center}
\caption{\label{fig:mad} Illustration of robot movements in a potential hierarchical strategy.}
\end{figure}

By introducing additional hierarchies, Strategy \ref{algs4} can be easily extended to a multi-hierarchy decentralized strategy. Depending on how the subdivisions are made, many such strategies are possible. For example, using $h \ge 2$ hierarchies with each hierarchy $i$ having $4^{i-1}$ small squares, we get a ``quad-merging'' strategy as illustrated in Fig. \ref{fig:mad}, in which up to four representatives in four adjacent squares meet to decide a local assignment of the robots in these squares at a given hierarchy level. 

Although these suboptimal strategies vary in detail, they can be easily analyzed with Theorem \ref{t:hier}. For example, we look at an extension to Strategy \ref{algs4} with three hierarchies; let us call this strategy Strategy \customlabel{algs5}{5}\ref{algs5}. After partitioning the bottom (or third) hierarchy to $m$ squares, the middle (or second) hierarchy is partitioned into $k = \sqrt{m}$ small squares. At either the third or the second hierarchy, local assignments are made, followed by applying the rendezvous strategy as given in Strategy \ref{alg:rend}. It is again straightforward to derive the following.

\begin{corollary} For Strategy \ref{algs5} (3-level Hierarchical Rendezvous), as $n \to \infty$, 
\begin{equation}\label{eq:dch5}
\mathbf{E}[D_n] \le C\sqrt{n\log n} + 2\sqrt{n\sqrt{m}}  + 4\sqrt{m} + 2. 
\end{equation}
\end{corollary}

{\bf Remark.} Again, $D_n/D_n^* = O(1)$ as $n \to \infty$ for a fixed $m$. Introducing more hierarchy levels extends the total completion time $T_n$, which is increased by approximately $2\sqrt{m}$. Thus, the total completion time of Strategy \ref{algs5} is also given by (\ref{eq:tn1}). Following similar analysis, the overall running time required by Strategy \ref{algs5} is $O(mt(n/m) + \sqrt{m}t(\sqrt{n}) + t(\sqrt{n\sqrt{m}}))$ given a centralized assignment algorithm that runs in $t(n)$ time. 

\subsection{Handling Arbitrary $r_{\mathrm{sense}}$}
Because there can be targets anywhere in $Q$, to carry out the algorithms stated in this section, each robot must be aware of all target locations. For this to happen for arbitrary $r_{\mathrm{sense}}$, $Q$ must be swept through in a worst scenario. To do this, we partition $Q$ into $\lceil 1/(2r_{\mathrm{sense}}) \rceil^2$ small squares and let a robot in the top-left small square ``zig-zag'' through $Q$ (i.e., following a Boustrophedon path \cite{Cho00}) until it covers the bottom side of $Q$. If there is no robot in the top-left small square, then a robot in a square along the Boustrophedon path is used; implicit timing can be used to determine this. Once the end of the path is reached, the robot then reverses its course until it gets back to the top-left small square. At this point, this robot is aware of all target locations. It can then repeat a similar path to communicate that information to all other robots. This procedure ensures that all robots are aware of all target locations. The total distance cost of the procedure is about $2\lceil 1/(2r_{\mathrm{sense}}) \rceil + \lceil 1/(2r_{\mathrm{comm}})\rceil$. Taking this penalty, which does not depend on $n$ and therefore has no impact on the asymptotic optimality, we can then effectively assume $r_{\mathrm{sense}} \ge \sqrt{2}$.

\section{Simulation Studies}\label{sec:exp}

\subsection{Number of Required Robots for a Connected $G(0)$}

\begin{figure}[htp]
\begin{center}
    \includegraphics[width=3in]{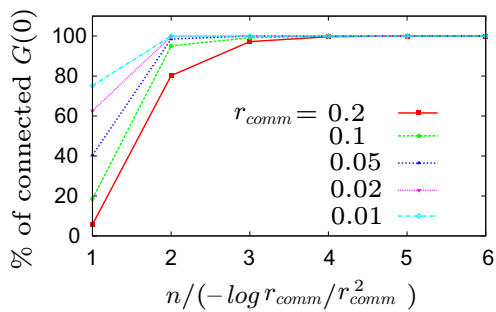}
\end{center}
\caption{\label{fig:conng} Effects of $n$ on the connectivity of $G(0)$ for different values of $r_{\mathrm{comm}}$.}
\end{figure}

In this subsection, we show a result of simulation to verify our theoretical findings in Section \ref{sec:bound}. Since the bounds over $r_{\mathrm{comm}}$ and $r_{\mathrm{sense}}$ are similar, we focus on $r_{\mathrm{comm}}$ and verify the requirement for the connectivity of $G(0)$ for several $r_{\mathrm{comm}}$'s ranging from $0.01$ to $0.2$. For each fixed $r_{\mathrm{comm}}$, various numbers of robots are used starting from $n = -\log r_{\mathrm{comm}}/{r_{\mathrm{comm}}^2}$ (the number of robots goes as high as $3 \times 10^5$ for the case of $r_{\mathrm{comm}} = 0.01$). 1000 trials were run for each fixed combination of $r_{\mathrm{comm}}$ and $n$. The percentage of the runs with a connected $G(0)$ is reported in Fig. \ref{fig:conng}. The simulation suggests that the bounds on $n$ from Theorem \ref{t:comm} are fairly tight. 

\begin{table}[htp]
\begin{center}
	 \caption{\label{table:comm}Comparison between (\ref{eq:rgg}) and (\ref{eq:comm2})}
	 \begin{tabular}{cccccc}
   \hline\hline
	 \multirow{2}*{prob.} & \multicolumn{5}{c}{$r_{\mathrm{comm}}$} \\
	 \cline{2-6}
& 0.2	&	0.1	&	0.05	&	0.02	&	0.01 \\ \hline
0.1 & 0.001, 0.82	&	0.001, 0.96 	&	0.001, 0.99	&	0.001, 1	&	0.003, 1 \\ \hline
0.5 & 0.007, 0.92	&	0.006, 0.98	&	0.027, 0.99	&	0.064, 1	&	0.081, 1 \\ \hline
0.9 & 0.2, 0.99	&	0.31, 1	&	0.381, 1	&	0.477, 1	&	0.502, 1 \\ \hline
0.99 & 0.702, 1	&	0.742, 1	&	0.794, 1	&	0.834, 1	&	0.855, 1 \\ 
	 \hline\hline
	 \end{tabular}
\end{center}
\end{table}

To compare to (\ref{eq:rgg}), which also allows for estimation of $n$ in terms of $r_{\mathrm{comm}}$ with a specified probability for obtaining a connected $G(0)$, we computed $n$ based on (\ref{eq:rgg}) and (\ref{eq:comm2}) for a range of $r_{\mathrm{comm}}$-probability pairs. We then use these $n$'s to estimate the actual probability of having a connected $G(0)$. We list the result in Table \ref{table:comm}. Each main entry of the table has two probability numbers separated by a comma, obtained using (\ref{eq:rgg}) and (\ref{eq:comm2}), respectively. As we can see, (\ref{eq:rgg}) gives underestimates (due to its asymptotic nature) and cannot be used to provide probabilistic guarantees. On the other hand, (\ref{eq:comm2}) provides overestimates that guarantee the desired probability. 

\subsection{Performance of Near Optimal Strategies}
Next, we simulate Strategies \ref{alg:hier}-\ref{algs5} and evaluate $D_n$, $T_n$, and running time for these strategies over various values of $n$ and $r_{\mathrm{comm}}$, assuming $r_{\mathrm{sense}} \ge \sqrt{2}$. Due to our choice of $k = \sqrt{m}$ in Strategy \ref{algs5}, we pick specific $r_{\mathrm{comm}}$'s so that $m = \lceil \sqrt{2}/r_{\mathrm{comm}} \rceil$ is always a perfect square. These values are $r_{\mathrm{comm}} = 0.16, 0.09, 0.057$, and $0.04$, which correspond to $m = 81, 256, 625$, and $1296$, respectively. The number of robots used in each simulation ranges from $100$ to $10000$. For each $n$, $10$ assignment problem instances are randomly generated. These problem instances are then used to test all strategies. We test Strategy \ref{alg:hier} using the same (two-hierarchy and three-hierarchy) partitions that are used with Strategies \ref{algs4} and \ref{algs5}.  

\begin{figure}[htp]
\begin{center}
    \includegraphics[width=3in]{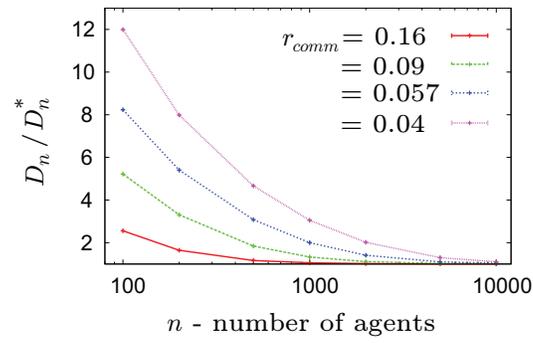}
\end{center}
\caption{\label{fig:s1} Distance optimality of Strategy \ref{alg:rend} over varying $n$ and $r_{\mathrm{comm}}$.}
\end{figure}

\subsubsection*{Distance optimality} The ratios $D_n/D_n^*$ for Strategy \ref{alg:rend} over different $n$ and $r_{\mathrm{comm}}$ are plotted in Fig. \ref{fig:s1}. We observe that the overhead for establishing global communication among the robots becomes insignificant as $n$ increases, driving $D_n/D_n^*$ to close to one. 

\begin{figure}[htp]
\begin{center}
    \includegraphics[width=3in]{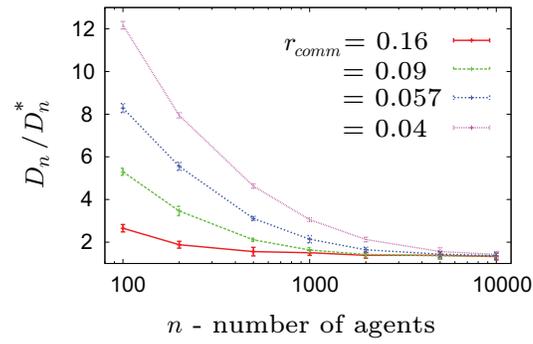}
\end{center}
\caption{\label{fig:h2} Distance optimality of Strategy \ref{algs4} over varying $n$ and $r_{\mathrm{comm}}$.}
\end{figure}

\begin{figure}[htp]
\begin{center}
    \includegraphics[width=3in]{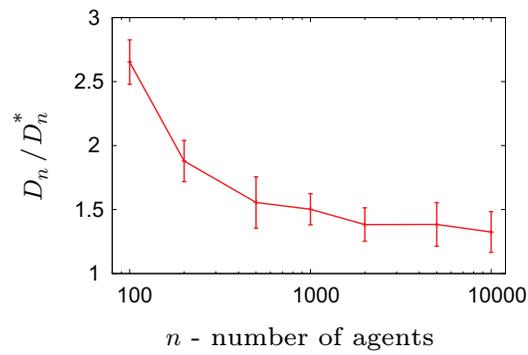}
\end{center}
\caption{\label{fig:m9} The effect of varying $n$ on the distance optimality of Strategy \ref{algs4} with $r_{\mathrm{comm}} = 0.16$ ($m = 81$).}
\end{figure}

For Strategy \ref{algs4}, the ratios were plotted similarly in Fig. \ref{fig:h2} but with (small) error bars. The error bars display the standard deviation over the $10$ runs (we omitted these from a figure, such as Fig. \ref{fig:s1}, when they are too small to see). They can be better seen in Fig. \ref{fig:m9}, which is a zoomed-in version of the $r_{\mathrm{comm}} = 0.16$ line from Fig. \ref{fig:h2}. The similarities between Fig.~\ref{fig:s1} and Fig.~\ref{fig:h2} for small $n$ are not surprising since both strategies spend most of their effort (distance traveled) in establishing communication. As this extra communication cost diminishes as $n$ grows, the actual assignment cost dominates. Strategy \ref{alg:rend}, with assignment being done in a centralized manner, becomes better than the decentralized Strategy \ref{algs4}. 
\begin{figure}[htp]
\begin{center}
    \includegraphics[width=3in]{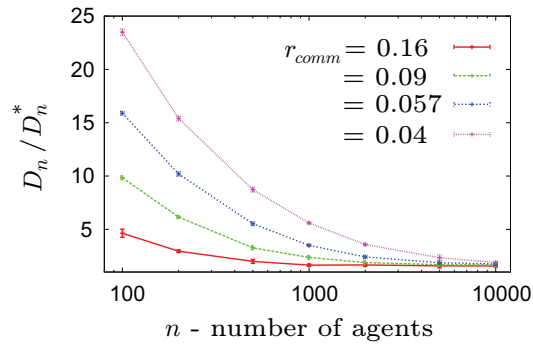}
\end{center}
\caption{\label{fig:h3} Distance optimality of Strategy \ref{algs5} over varying $n$ and $r_{\mathrm{comm}}$.}
\end{figure}

As expected, for a fixed $r_{\mathrm{comm}}$, $D_n/D_n^*$ decreases as $n$ increases. For $n = 10000$, the approximation ratios for our choices of $r_{\mathrm{comm}}$ are around 1.4 (due to the slow growing nature of $D_n^* \sim \sqrt{n\log n}$; fixing any $r_{\mathrm{comm}}$, this ratio should be close to one for large $n$). On the other hand, for a fixed $n$, as the partition of the unit square $Q$ gets finer,  $D_n/D_n^*$ increases, implying that decreasing the communication radius has a negative effect on distance optimality. We observe similar results on the distance optimality of Strategy \ref{algs5} (see Fig. \ref{fig:h3}). 

\begin{figure}[htp]
\begin{center}
    \includegraphics[width=3in]{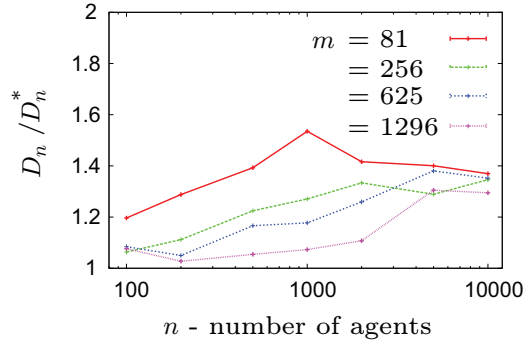}
\end{center}
\caption{\label{fig:h2diff} The assignment cost of a two-level ``pure'' hierarchical strategy.}
\end{figure}

\begin{figure}[htp]
\begin{center}
    \includegraphics[width=3in]{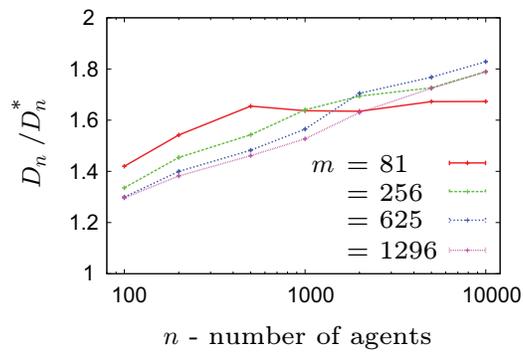}
\end{center}
\caption{\label{fig:h3diff} The assignment cost of a three-level ``pure'' hierarchical strategy.}
\end{figure}

If we remove the rendezvous part from Strategies \ref{algs4} and \ref{algs5}, they become similar to Strategy \ref{alg:hier}. The distance optimality performance of these two particular versions of Strategy \ref{alg:hier} is shown in Fig. \ref{fig:h2diff} and Fig. \ref{fig:h3diff}, respectively. For all partitions made ($m = 81, 256, 625, 1296$), $D_n/D_n^*$ ratios of less than two are achieved and can go as low as 1.06, showing that hierarchical strategies can provide very good optimality guarantees.

\subsubsection*{Computational performance} We list the running time, in seconds, for Strategies \ref{alg:rend}-\ref{algs5} in Table \ref{table:ct}. The standard $O(n^3)$ Hungarian method is used as the baseline assignment algorithm. Each main entry of the table lists three numbers corresponding to the running time of Strategies \ref{alg:rend}, \ref{algs4}, and \ref{algs5}, respectively, for the given combination of $r_{\mathrm{comm}}$ and $n$. Note that any version of Strategy \ref{alg:hier} has the same amount of computation as a corresponding rendezvous-based strategy. As expected, a hierarchical assignment greatly reduces the computation time, often by a factor over $10^3$. The computation was performed on a Intel Core-i7 3970K PC under a 8GB Java virtual machine.

\begin{table}[htp]
\begin{center}
	 \caption{\label{table:ct}Running time for Strategies \ref{alg:rend}-\ref{algs5}}
	 \begin{tabular}{ccccc}
   \hline\hline
	 \multirow{2}*{\# of robots, $n$} & \multicolumn{4}{c}{$r_{\mathrm{comm}} (m)$} \\
	 \cline{2-5}
	  & 0.16 (81) & 0.09 (256) & 0.057 (625) & 0.04 (1296) \\
	 \hline
	 100 & \begin{tabular}{c} 0.007 \\ 0.001 \\ 0.001 \end{tabular} 
	 & \begin{tabular}{c} 0.007  \\ 0.002 \\ 0.0001 \end{tabular} 
	 & \begin{tabular}{c} 0.007  \\ 0.002 \\ 0.0004 \end{tabular}
	 & \begin{tabular}{c} 0.007 \\ 0.003 \\ 0.0004 \end{tabular} 
	 \\
	 \hline
	 200 &	 \begin{tabular}{c} 0.02 \\ 0.001  \\ 0.0001  \end{tabular}& 
	 \begin{tabular}{c} 0.02  \\  0.005 \\ 0.0003 \end{tabular}& 
	 \begin{tabular}{c} 0.02 \\ 0.01 \\ 0.0004 \end{tabular}& 
	 \begin{tabular}{c} 0.02 \\ 0.02 \\ 0.0006 \end{tabular} \\
	 \hline
	 500 &	 \begin{tabular}{c} 0.34 \\ 0.005 \\ 0.0006  \end{tabular}& 
	 \begin{tabular}{c} 0.34 \\ 0.02  \\ 0.001 \end{tabular}& 
	 \begin{tabular}{c} 0.34 \\ 0.07 \\ 0.002 \end{tabular}& 
	 \begin{tabular}{c} 0.34 \\ 0.14 \\ 0.003 \end{tabular} \\
	 \hline
	 1000 &	 \begin{tabular}{c} 2.76 \\ 0.015 \\ 0.002  \end{tabular}& 
	 \begin{tabular}{c} 2.76 \\ 0.07  \\ 0.003 \end{tabular}& 
	 \begin{tabular}{c} 2.76 \\ 0.22 \\ 0.003 \end{tabular}& 
	 \begin{tabular}{c} 2.76 \\ 0.54 \\ 0.006  \end{tabular} \\
	 \hline
	 2000 &	 \begin{tabular}{c} 22.3 \\ 0.05 \\ 0.009 \end{tabular}& 
	 \begin{tabular}{c} 22.3 \\ 0.20 \\ 0.006 \end{tabular}& 
	 \begin{tabular}{c} 22.3 \\ 0.70 \\ 0.011 \end{tabular}& 
	 \begin{tabular}{c} 22.3 \\ 1.90 \\ 0.015 \end{tabular} \\
	 \hline
	 5000 &	 \begin{tabular}{c} 345 \\ 0.02 \\ 0.069 \end{tabular}& 
	 \begin{tabular}{c} 345 \\ 0.78 \\ 0.032 \end{tabular}& 
	 \begin{tabular}{c} 345 \\ 2.84 \\ 0.043 \end{tabular}& 
	 \begin{tabular}{c} 345 \\ 8.28 \\ 0.058 \end{tabular} \\
	 \hline
	 10000 &	 \begin{tabular}{c} 2756  \\ 0.83 \\ 0.43  \end{tabular}& 
	 \begin{tabular}{c} 2756 \\ 2.32 \\ 0.11 \end{tabular}& 
	 \begin{tabular}{c} 2756 \\ 8.35 \\ 0.11 \end{tabular}& 
	 \begin{tabular}{c} 2756 \\ 24.4 \\ 0.14 \end{tabular} \\
	 \hline\hline
	 \end{tabular}
\end{center}
\end{table}

\subsubsection*{Time optimality} 
\begin{figure}[htp]
\begin{center}
\includegraphics[width=3in]{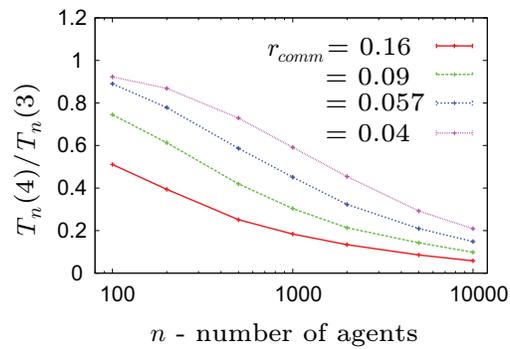} 
\end{center}
\caption{\label{fig:tn1} Ratio of total completion time between Strategies \ref{alg:rend} and \ref{algs4}.}
\end{figure}

\begin{figure}[htp]
\begin{center}
    \includegraphics[width=3in]{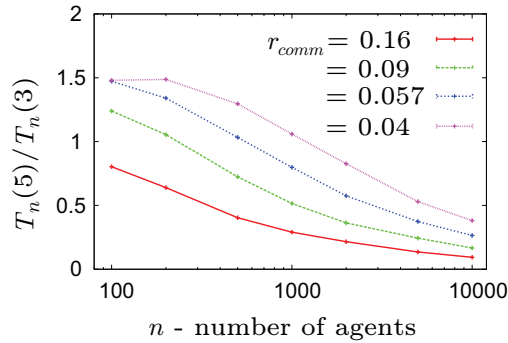}
\end{center}
\caption{\label{fig:tn2} Ratio of total completion time between Strategies \ref{alg:rend} and \ref{algs5}.}
\end{figure}

Since Strategies \ref{alg:rend}-\ref{algs5} sacrifice distance (and therefore, time) to compensate for limited communication, we do not expect the total completion time $T_n$ of these strategies to match $T_n^*$ closely. For example, in (\ref{eq:tn1}), although $T_n \to T_n^*$ as $n \to \infty$ for fixed $m = \lceil \sqrt{2}/r_{\mathrm{comm}} \rceil^2$, it requires a very large $n$ for $\sqrt{\log n}$ to dominate $\sqrt{m}$. Thus, we only compare $T_n$ among Strategies \ref{alg:rend}-\ref{algs5}. Using $T_n(i)$ to denote the $T_n$ for Strategy $i$, $T_n($\ref{algs4}$)/T_n($\ref{alg:rend}$)$ and $T_n($\ref{algs5}$)/T_n($\ref{alg:rend}$)$ are plotted in Figures \ref{fig:tn1}-\ref{fig:tn2}. As $n$ increases, Strategies \ref{algs4} and \ref{algs5} both take much less total completion time on average. 

\section{Conclusion and Discussions}\label{sec:con}

Focusing on the distance optimality for the target assignment problem in a robotic network setting, we have characterized a necessary and sufficient condition under which optimality can be achieved. We also provided a direct formula for computing the number of robots sufficient for probabilistically guaranteeing such an optimal solution. Then, we took a different angle; we looked at suboptimal strategies and their asymptotic performance as the number of robots goes to infinity. We showed that these strategies yield a constant approximation ratio when compared with the true distance optimal solution. Many of these decentralized strategies also provide computational advantages over a centralized one. 

We conclude the paper by discussing our choice on certain elements that can be generalized in a future work.

\subsubsection*{Equal number of initial and target locations} In the problem statement we assume that $|X^0| = |Y^0|$. If $|X^0| > |Y^0|$, some robots do not need to move and if $|X^0| < |Y^0|$, some robots may need to reach multiple targets, assuming that the main goal is to serve the targets. Our result readily generalizes to the case in which $|X^0|/|Y^0|$ is close to $1$. When $|X^0| \gg |Y^0|$, it is likely that for a $y_i \in Y^0$, there is a unique $x_i \in X^0$ that is closest to $y_i$ \cite{SmiBul09}. Moreover, for two different $y_i, y_j$, $x_i \ne x_j$. The spatial assignment problem then degenerates to finding the nearest robot for each $y \in Y^0$. When $|X^0| \ll |Y^0|$, the problem becomes a multiple salesmen version of the traveling salesman problem (we have a standard traveling salesman problem when $|X^0| = 1$), which is an NP-hard problem. It remains an interesting open question to investigate the middle ground, {\em i.e.}, $|X^0| = C|Y^0|$ for some constant $C$ (for example $C \in [0.1, 10]$). 

\subsubsection*{Distribution of initial and target locations} Although it is beyond the scope of this paper, it would be interesting to establish a lower bound on the optimal assignment distance for  arbitrary probability measures. Also, it would be interesting to investigate the case in which the robots and the targets assume different distributions. Another important aspect not covered in this paper is the issue of targets distributed somewhat randomly over time. 

\subsubsection*{Minimizing over other powers of the $2$-norm} On the side of optimality measures, we note that Theorem~\ref{t:bound} generalizes to arbitrary powers of the Euclidean $2$-norm \cite{AjtKomTus84}. That is, for 
\begin{equation}\label{eq:dnp}
D_{n,p}^* := \min_{\sigma}\sum_{i = 1}^n \Vert x_i^0 - y_{\sigma(i)}^0 \Vert_2^p, 
\end{equation}
it holds true that 
\begin{equation}\label{eq:dnpo}
D_{n,p}^* \sim n(\log n/n)^{p/2}. 
\end{equation}
Theorem~\ref{t:bound} corresponds to the special case of $p = 1$. As $p \to \infty$, (\ref{eq:dnp}) minimizes the longest distance traveled by any robot. This is true because for fixed $X^0$, $Y^0$, and a sufficiently large $p$, the largest $\Vert x_i^0 - y_{\sigma(i)}^0 \Vert_2^p$ becomes the dominating term in the sum $\sum_{i = 1}^n \Vert x_i^0 - y_{\sigma(i)}^0 \Vert_2^p$. Although we restrict our attention to $p = 1$ in this paper, our results readily extend to other values of $p$ ({\em i.e.}, other optimality criteria) with (\ref{eq:dnpo}). Note that this means the $D_n$ definition given by~\eqref{eq:sla} needs to be updated accordingly to an appropriately defined $D_{n,p}$.

\section*{Acknowledgements} We thank anonymous reviewers, S. Har-Peled, and A. Nayyeri for their constructive comments.

\bibliographystyle{IEEETranN}
\bibliography{reference}

\end{document}